\documentclass[review]{fcs}

\usepackage{booktabs}
\usepackage{color}
\usepackage{tikz}
\usepackage{xcolor}
\usepackage{xspace}
\usepackage{booktabs}
\usepackage{multirow}
\usepackage{makecell}
\usepackage[edges]{forest}
\definecolor{hidden-draw}{RGB}{20,68,106}
\definecolor{hidden-pink}{RGB}{255,245,247}
\usepackage[numbers,sort]{natbib}

\title{Large Language Models for Generative Information Extraction: A Survey}
\author[1,2]{Derong Xu\dag}
\author[1]{Wei Chen\dag}
\author[1]{Wenjun Peng}
\author[1,2]{Chao Zhang}
\author*[1]{Tong Xu}
\author*[2]{Xiangyu Zhao}
\author*[3]{Xian Wu}
\author[3]{Yefeng Zheng}
\author[4]{Yang Wang}
\author*[1]{Enhong Chen}

\address[1]{State Key Laboratory of Cognitive Intelligence \& University of Science and Technology of China, Hefei 230000, China}
\address[2]{Department of Data Science, City University of Hong Kong, Hongkong 999077, China} 
\address[3]{Jarvis Research Center, Tencent YouTu Lab, Beijing 100029, China}
\address[4]{Anhui Conch Information Technology Engineering Co., Ltd., Wuhu 241000, China}

\fcssetup{
  received       = {month dd, yyyy},
  accepted       = {month dd, yyyy},
  corr-email     = {\{derongxu,chenweicw,pengwj,zclfe00\}@mail.ustc.edu.cn, \{tongxu,cheneh\}@ustc.edu.cn,xianzhao@cityu.edu.hk, \{kevinxwu, yefengzheng\}@tencent.com, wangyang@chinaconch.com\\ \dag Equal Contribution.  The article has been accepted by Frontiers of Computer Science (FCS), with the DOI: {10.1007/s11704-024-40555-y}. You can cite the FCS version.},
}
\begin{abstract}
Information extraction (IE) aims to extract structural knowledge from plain natural language texts. Recently, generative Large Language Models (LLMs) have demonstrated remarkable capabilities in text understanding and generation. As a result, numerous works have been proposed to integrate LLMs for IE tasks based on a generative paradigm. To conduct a comprehensive systematic review and exploration of LLM efforts for IE tasks, in this study, we survey the most recent advancements in this field. We first present an extensive overview by categorizing these works in terms of various IE subtasks and techniques, and then we empirically analyze the most advanced methods and discover the emerging trend of IE tasks with LLMs. Based on a thorough review conducted, we identify several insights in technique and promising research directions that deserve further exploration in future studies. We maintain a public repository and consistently update related works and resources on GitHub (\href{https://github.com/quqxui/Awesome-LLM4IE-Papers}{LLM4IE repository}).
\end{abstract}

\begin{document}

\section{Introduction}
Information Extraction (IE) is a crucial domain in natural language processing (NLP) that converts plain text into structured knowledge (e.g., entities, relations, and events), and serves as a foundational requirement for a wide range of downstream tasks, such as knowledge graph construction \cite{zhong2023comprehensive}, knowledge reasoning \cite{fu2019collaborative} and question answering \cite{srihari1999information}. Typical IE tasks consist of Named Entity Recognition (NER), Relation Extraction (RE) and Event Extraction (EE) \cite{uie,instructuie,code4uie,zhong2023contextualized}. 
However, performing IE tasks is inherently challenging. This is because IE involves extracting information from various sources and dealing with complex and ever-changing domain requirements \cite{zhou2022survey}. Unlike traditional NLP tasks, IE encompasses a broad spectrum of objectives such as entity extraction, relationship extraction, and more. In IE, the extraction targets exhibit intricate structures where entities are presented as span structures (string structures) and relationships are represented as triple structures \cite{uie}.
Additionally, in order to effectively handle different information extraction tasks, it is necessary to employ multiple independent models. These models are trained separately for each specific task, without sharing any resources. However, this approach comes with a drawback: managing a large number of information extraction models becomes costly in terms of the resources needed for construction and training, like annotated corpora.

The emergence of large language models (LLMs), such as GPT-4 \cite{gpt4}, has significantly advanced the field of NLP, due to their extraordinary capabilities in text understanding and generation \cite{qi2024unimel,peng2024large,zhang2024notellm2}.
Pretraining LLMs using auto-regressive prediction allows them to capture the inherent patterns and semantic knowledge within text corpora \cite{liu2023pre,lyu2024crud,lyu2024retrieve,jia-etal-2024-mill,10.1145/3589335.3648321,fu2023unified,jia2024g3}. This enhances LLMs with the capability to perform zero-shot and few-shot learning, enabling them to model various tasks consistently and serving as tools for data augmentation \cite{zhang2024notellm, wang2024context, zhu2024fastmem}. Furthermore, LLMs can serve as intelligent agents for complex task planning and execution, utilizing memory retrieval and various tools to enhance efficiency and successfully accomplish tasks \cite{wang2024survey, guan2024enhancing, huang2024qdmr,fu2024video,li2023agent4ranking}.
Therefore, there has been a recent surge of interest in generative IE methods \cite{qi2023preserving} that adopt LLMs to generate structural information rather than extracting structural information from plain text. These methods have been proven to be more practical in real-world scenarios compared to discriminated methods \cite{chen2023heproto,usm}, as they can handle schemas containing millions of entities without significant performance degradation \cite{genie}. 

\begin{figure*}[t]
\begin{center}
\includegraphics[scale=0.8]{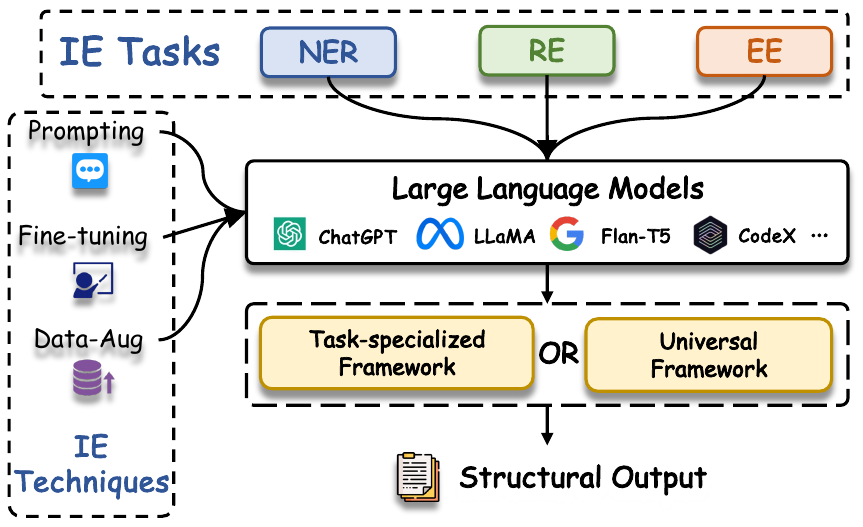} 
\caption{LLMs have been extensively explored for generative IE. These studies encompass various IE techniques, specialized frameworks designed for a single subtask, and universal frameworks capable of addressing multiple subtasks simultaneously.}
\label{fig:intro}
\end{center}
\end{figure*}

On the one hand, LLMs have attracted significant attention from researchers in exploring their potentials for various scenarios and tasks of IE. 
In addition to excelling in individual IE tasks, LLMs possess a remarkable ability to effectively model all IE tasks in a universal format. 
This is conducted by capturing inter-task dependencies with instructive prompts, and achieves consistent performance \cite{uie,gollie,paolini2021structured,instructuie,giellm,fei2022lasuie,codeie}.
On the other hand, recent works have shown the outstanding generalization of LLMs to not only learn from IE training data through fine-tuning \cite{paolini2021structured,yan2021unified,tempgen,rebel,paolini2021structured}, but also extract information in few-shot and even zero-shot scenarios relying solely on in-context examples or instructions \cite{chatie,code4struct,gpt-ner,promptner,xu2023unleash}. 

However, existing surveys \cite{nasar2021named,zhou2022survey,ye2022generative} do not provide a comprehensive exploration of these areas for the above two groups of research works: 1) universal frameworks that cater to multiple tasks and 2) cutting-edge learning techniques for scenarios with limited training data. The community urgently needs a more in-depth analysis of how LLM can be more appropriately applied to IE tasks to improve the performance of the IE field. This is because there are still challenges and issues in applying LLM to IE in terms of learning and understanding knowledge \cite{foppiano2024mining}. These challenges include the misalignment between natural language output and structured form \cite{code4uie}, hallucination problem in LLMs \cite{liu2024survey}, contextual dependence, high computational resource requirements \cite{sahoo2024systematic}, difficulties in updating internal knowledge \cite{xu2024editing}, etc.

In this survey, we provide a comprehensive exploration of LLMs for generative IE, as illustrated in Fig. \ref{fig:intro}. To achieve this, we categorize existing methods mainly using two taxonomies: (1) a taxonomy of numerous IE subtasks, which aims to classify different types of information that can be extracted individually or uniformly, and (2) a taxonomy of IE techniques, which categorizes various novel approaches that utilize LLMs for generative IE, particularly on low-resource scenarios. In addition, we present a comprehensive review of studies that specifically focus on the application of IE techniques in various domains. And we discuss studies that aim to evaluate and analyze the performance of LLMs for IE.
According to the above division, we construct a taxonomy of related studies as shown in Fig. \ref{fig_taxonomy}.
We also compare several representative methods to gain deeper understanding of their potentials and limitations, and provide insightful analysis on future directions. 
To the best of our knowledge, this is the first survey on generative IE with LLMs. 

The remaining part of this survey is organized as follows: 
We first introduce the definition of generative IE and target of all subtasks in Section \ref{sec_prelimi}. Then, in Section \ref{sec_task}, we introduce representative models for each task and universal IE, and compare their performance.
In Section \ref{sec:paradigm}, we summarize different learning techniques of LLMs for IE. Additionally, we introduce works proposed for special domains in Section \ref{sec:domain}, and present recent studies that evaluate and analyze the abilities of LLMs on IE tasks in Section \ref{sec:eval}. Finally, we propose potential research directions for future studies in Section \ref{sec:future}. In Section \ref{sec:Benchmarks}, we provide a comprehensive summary of the most commonly used LLMs and datasets statistics, as reference for researchers.

\tikzstyle{my-box}=[
    rectangle,
    draw=hidden-draw,
    rounded corners,
    align=left,
    text opacity=1,
    minimum height=1.5em,
    minimum width=5em,
    inner sep=2pt,
    fill opacity=.8,
    line width=0.8pt,
]

\tikzstyle{leaf-head}=[my-box, minimum height=1.5em,
    draw=gray!80, 
    fill=gray!15,  
    text=black, font=\normalsize,
    inner xsep=2pt,
    inner ysep=4pt,
    line width=0.8pt,
]

\tikzstyle{leaf-task}=[my-box, minimum height=1.5em,
    draw=red!70, 
    fill=red!15,  
    text=black, font=\normalsize,
    inner xsep=2pt,
    inner ysep=4pt,
    line width=0.8pt,
]

\tikzstyle{leaf-paradigms}=[my-box, minimum height=1.5em,
    draw=cyan!70, 
    fill=cyan!15,  
    text=black, font=\normalsize,
    inner xsep=2pt,
    inner ysep=4pt,
    line width=0.8pt,
]
\tikzstyle{leaf-others}=[my-box, minimum height=1.5em,
    draw=orange!80, 
    fill=orange!15,  
    text=black, font=\normalsize,
    inner xsep=2pt,
    inner ysep=4pt,
    line width=0.8pt,
]

\tikzstyle{modelnode-task}=[my-box, minimum height=1.5em,
    draw=red!80, 
    fill=hidden-pink!30,  
    text=black, font=\footnotesize,
    inner xsep=2pt,
    inner ysep=4pt,
    line width=0.8pt,
]

\tikzstyle{modelnode-paradigms}=[my-box, minimum height=1.5em,
    draw=cyan!100, 
    fill=hidden-pink!30,  
    text=black, font=\footnotesize,
    inner xsep=2pt,
    inner ysep=4pt,
    line width=0.8pt,
]

\tikzstyle{modelnode-others}=[my-box, minimum height=1.5em,
    draw=orange!100, 
    fill=hidden-pink!30,  
    text=black, font=\footnotesize,
    inner xsep=2pt,
    inner ysep=4pt,
    line width=0.8pt,
]

\begin{figure*}[htbp]
    \centering
    \resizebox{1\textwidth}{!}{
        \begin{forest}
            forked edges,
            for tree={
                grow=east,
                reversed=true,
                anchor=base west,
                parent anchor=east,
                child anchor=west,
                base=left,
                font=\normalsize,
                rectangle,
                draw=hidden-draw,
                rounded corners,
                align=left,
                minimum width=1em,
                edge+={darkgray, line width=1pt},
                s sep=3pt,
                inner xsep=0pt,
                inner ysep=3pt,
                line width=0.8pt,
                ver/.style={rotate=90, child anchor=north, parent anchor=south, anchor=center},
            }, 
            [
                LLMs for Generative Information Extraction,leaf-head, ver
                [
                     Information \\Extraction \\Tasks (\S\ref{sec_task}), leaf-task,text width=5em
                    [
                        Named Entity \\ Recognition \\ (\S\ref{sec_ner}), leaf-task, text width=5.5em
                        [
                            Typing, leaf-task, text width=5.5em
                            [GET~\cite{get}{, }CASENT~\cite{casent}, modelnode-task, text width=45em]
                        ]
                        [
                            Identification \\\& Typing, leaf-task, text width=5.5em
                            [Yan et al.~\cite{yan2021unified}{, }TEMPGEN~\cite{tempgen}{, }Cui et al.~\cite{cui2021template}{, }Zhang et al.~\cite{zhang2022bias}{, }Wang et al.~\cite{instructionner}{, }Xia et al.~\cite{xia2023debiasing}{, }Cai et al.~\cite{cai2023context}{, }EnTDA~\cite{entda}{, }\\Amalvy et al.~\cite{amalvy2023learning}{, }GPT-NER~\cite{gpt-ner}{, }Cp-NER~\cite{cp-ner}{, }LLMaAA~\cite{llmaaa}{, }PromptNER~\cite{promptner}{, }Ma et al.~\cite{ma2023large}{, }Xie et al.~\cite{xie2024self}{, }UniNER~\cite{universalner}{, }\\NAG-NER~\cite{nag-ner}{, }Su et al.~\cite{SuY23}{, }GNER~\cite{gner}{, }NuNER~\cite{nuner}{, }MetaNER~\cite{metaner}{, }LinkNER~\cite{linkner}{, }SLCoLM~\cite{slcolm}{, }Popovivc et al.~\cite{popovivc2024embedded}{, }\\ProgGen~\cite{proggen}{, }C-ICL~\cite{c-icl}{, }Keloth et al.~\cite{keloth2024advancing}{, }VerifiNER~\cite{verifiner}{, }Li et al.~\cite{li2024simple}{, }Oliveira et al.~\cite{oliveira2024combining}{, }Lu et al.~\cite{padellm-ner}{, }Bolucu et al.~\cite{bolucu2023impact}{, }\\Liu et al.~\cite{liu2023improving}{, }ConsistNER~\cite{consistner}{, }Naguib et al.~\cite{naguib2024few}{, }GLiNER~\cite{gliner}{, }Munnang et al.~\cite{munnangi2024fly}{, }Zhang et al.~\cite{zhang2024context}{, }LTNER~\cite{yan2024ltner}{, }\\ToNER~\cite{jiang2024toner}{, }Nunes et al.~\cite{nunesout}{, }Hou et al.~\cite{hou2024knowledge}{, }RT~\cite{li2024rt}{, }Jiang et al.~\cite{jiang2024p}{, }VANER~\cite{biana2024vaner}{, }RiVEG~\cite{riveg}{, }LLM-DA~\cite{ye2024llm}{, }etc.
                            , modelnode-task, text width=45em]
                        ]
                    ]
                    [
                        Relation \\ Extraction \\ (\S\ref{sec_re}), leaf-task, text width=5.5em
                        [
                            Classification, leaf-task, text width=5.5em
                            [REBEL~\cite{rebel}{, }Li et al.~\cite{li2023revisiting}{, }GL~\cite{gl}{, }Xu et al.~\cite{xu2023unleash}{, }QA4RE~\cite{qa4re}{, }LLMaAA~\cite{llmaaa}{, }GPT-RE~\cite{gpt-re}{, }Ma et al.~\cite{ma2023large}{, }\\STAR~\cite{star}{, }AugURE~\cite{augure}{, }RELA~\cite{rela}{, }DORE~\cite{dore}{, }CoT-ER~\cite{cot-er}{, }REPAL~\cite{repal}{, }Qi et al.~\cite{qi2023mastering}{, }Li et al.~\cite{li2024meta}{, }\\Otto et al.~\cite{otto2024enhancing}{, }Shi et al.~\cite{shi2024cre}{, }Li et al.~\cite{li2024recall}{, }Li et al.~\cite{li2024empirical}{, }RAG4RE~\cite{efeoglu2024retrieval}{, }Li et al.~\cite{li2024relation},
                            modelnode-task, text width=45em]
                        ]
                        [
                            Triplet, leaf-task, text width=5.5em
                            [TEMPGEN~\cite{tempgen}{, }Fan et al.~\cite{FanH23}{, }Kwak et al.~\cite{KwakJFBMS23}{, }DocRTE~\cite{docrte}{, }REPLM~\cite{replm}{, }AutoRE~\cite{autore}{, }ERA-COT~\cite{era-cot}{, }\\Li et al.~\cite{li2024simple}{, }I2CL~\cite{i2cl}{, }SLCoLM~\cite{slcolm}{, }Qi et al.~\cite{qi2023mastering}{, }Liu et al.~\cite{liu2023improving}{, }Ding et al.~\cite{ding2024improving}{, }Li et al.~\cite{li2024meta}{, }Ni et al.~\cite{ni2023unified}, modelnode-task, text width=45em]
                        ]
                        [
                            Strict, leaf-task, text width=5.5em
                            [REBEL~\cite{rebel}{, }Zaratiana et al.~\cite{ZaratianaTHC24}{, }C-ICL~\cite{c-icl}{, }MetaIE~\cite{peng2024metaie}{, }Atuhurra et al.~\cite{atuhurra2024distilling}{, }CHisIEC~\cite{chisiec}, modelnode-task, text width=45em]
                        ]
                    ]
                    [
                        Event \\ Extraction \\ (\S\ref{sec_ee}), leaf-task, text width=5.5em
                        [
                            Detection, leaf-task, text width=5.5em
                            [Veyseh et al.~\cite{veyseh2021unleash}{, }DAFS~\cite{dafs}{, }Qi et al.~\cite{qi2023mastering}{, }Cai et al.~\cite{cai2024improving}{, }Ni et al.~\cite{ni2023unified}, modelnode-task,text width=45em]
                        ]
                        [
                            Argument \\ Extraction, leaf-task, text width=5.5em
                            [BART-Gen~\cite{bart-gen}{, }Text2Event~\cite{text2event}{, }ClarET~\cite{claret}{, }X-GEAR~\cite{huang2022multilingual}{, }PAIE~\cite{paie}{, }GTEE-DYNPREF~\cite{gtee-dynpref}{, }Cai et al.~\cite{cai2023monte}{, }\\Ma et al.~\cite{ma2023large}{, }GL~\cite{gl}{, }STAR~\cite{star}{, }Code4Struct~\cite{code4struct}{, }PGAD~\cite{pgad}{, }QGA-EE~\cite{qga-ee}{, }SPEAE~\cite{speae}{, }AMPERE~\cite{ampere}{, }\\KeyEE~\cite{duan2024keyee}{, }Lin et al.~\cite{LinZS23}{, }Liu et al.~\cite{liu2024beyond}{, }Peng et al.~\cite{peng2024metaie}{, }ULTRA~\cite{zhang2024ultra}{, }Sun et al.~\cite{sun2024leveraging}{, }Zhou et al.~\cite{zhou2023heuristics}, modelnode-task,text width=45em]
                        ]
                        [
                            Detection \& \\ Argument \\ Extraction, leaf-task, text width=5.5em
                            [BART-Gen~\cite{bart-gen}{, }Text2Event~\cite{text2event}{, }DEGREE~\cite{degree}{, }ClarET~\cite{claret}{, }GTEE-DYNPREF~\cite{gtee-dynpref}{, }Ma et al.~\cite{ma2023large}{, }GL~\cite{gl}{, }\\STAR~\cite{star}{, }Cai et al.~\cite{cai2023monte}{, }DemoSG~\cite{demosg}{, }Kwak et al.~\cite{KwakJFBMS23}{, }EventRL~\cite{eventrl}{, }Huang et al.~\cite{Huang23textee}, modelnode-task,text width=45em]
                        ]
                    ]
                    [
                        Universal \\ Information \\ Extraction \\ (\S\ref{sec_uniie}), leaf-task, text width=5.5em
                        [
                            NL-LLMs \\based, leaf-task, text width=5.5em
                            [TANL~\cite{paolini2021structured}{, }DEEPSTRUCT~\cite{deepstruct}{, }GenIE~\cite{genie}{, }UIE~\cite{uie}{, }LasUIE~\cite{fei2022lasuie}{, }ChatIE~\cite{chatie}{, }InstructUIE~\cite{instructuie}{, }GIELLM~\cite{giellm}{, }\\Set~\cite{set}{, }CollabKG~\cite{collabkg}{, }TechGPT-2~\cite{techgpt2}{, }YAYI-UIE~\cite{yayi-uie}{, }ChatUIE~\cite{chatuie}{, }IEPile~\cite{iepile}{, }Guo et al.~\cite{guo2024diluie}, modelnode-task, text width=45em]
                        ]
                        [
                            Code-LLMs \\based, leaf-task, text width=5.5em
                            [CODEIE~\cite{codeie}{, }CodeKGC~\cite{codekgc}{, }GoLLIE~\cite{gollie}{, }Code4UIE~\cite{code4uie}{, }KnowCoder~\cite{knowcoder}, modelnode-task, text width=45em]
                        ]
                    ]
                ]
                [
                    Information \\Extraction \\ Techniques \\ (\S\ref{sec:paradigm}), leaf-paradigms,text width=5em
                    [
                        Data Augme-\\ntation (\S\ref{sec:da}), leaf-paradigms, text width=5.5em
                        [
                            Data Annotation, leaf-paradigms,text width=9em
                            [Veyseh et al.~\cite{veyseh2021unleash}{, }LLMaAA~\cite{llmaaa}{, }AugURE~\cite{augure}{, }Xu et al.~\cite{xu2023unleash}{, }Li et al.~\cite{li2023semi}{, }Tang et al.~\cite{tang2023does}{, }REPAL~\cite{repal}{, }~\cite{oliveira2024combining}\\Meoni et al.~\cite{meoni2023large}{, }Evans et al.~\cite{evans2024astro}{, }MetaIE~\cite{peng2024metaie}{, }NuNER~\cite{nuner}{, }LLM-DA~\cite{ye2024llm}{, }Sun et al.~\cite{sun2024leveraging}{, }Naraki et al.~\cite{naraki2024augmenting}, modelnode-paradigms,text width=41.5em]
                        ]
                        [
                            Knowledge Retrieval, leaf-paradigms, text width=9em
                            [Amalvy et al.~\cite{amalvy2023learning}{, }\cite{chen2023chain}{, }PGIM~\cite{pgim}{, }DocRTE~\cite{docrte}{, }VerifiNER~\cite{verifiner}{, }RiVEG~\cite{riveg}, modelnode-paradigms,text width=41.5em]
                        ]
                        [
                            Inverse Generation, leaf-paradigms, text width=9em
                            [EnTDA~\cite{entda}{, }STAR~\cite{star}{, }QGA-EE~\cite{qga-ee}{, }SynthIE~\cite{synthie}{, }ProgGen~\cite{proggen}{, }Atuhurra et al.~\cite{atuhurra2024distilling}, modelnode-paradigms,text width=41.5em]
                        ]
                        [
                            Synthetic Datasets for \\Instruction-tuning , leaf-paradigms, text width=9em
                            [UniversalNER~\cite{universalner}{, }GLiNER~\cite{gliner}{, }GNER~\cite{gner}{, }Chen et al.~\cite{chen2023chain}, modelnode-paradigms,text width=41.5em]
                        ]
                    ]
                    [
                        Prompts Design \\(\S\ref{sec:promptdesign}), leaf-paradigms, text width=5.5em
                        [
                            Question Answer, leaf-paradigms, text width=9em
                            [QA4RE~\cite{qa4re}{, }ChatIE~\cite{chatie}{, }Hou et al.~\cite{hou2024knowledge}{, }Otto et al.~\cite{otto2024enhancing}{, }Li et al.~\cite{li2023revisiting},modelnode-paradigms, text width=41.5em]
                        ]
                        [
                            Chain of Thought, leaf-paradigms,text width=9em
                            [Chen et al.~\cite{chen2023chain}{, }PromptNER~\cite{promptner}{, }Wadhwa et al.~\cite{wadhwa2023revisiting}{, }Yuan et al.~\cite{yuan2023zero}{, }Bian et al.~\cite{bian2023inspire}{, }RT~\cite{li2024rt}, modelnode-paradigms,text width=41.5em]
                        ]
                        [
                            Self-Improvement, leaf-paradigms, text width=9em
                            [Xie et al.~~\cite{xie2024self}{, }ProgGen~\cite{proggen}{, }ULTRA~\cite{zhang2024ultra},modelnode-paradigms, text width=41.5em]
                        ]
                    ]
                    [
                        Zero-shot \\ (\S\ref{sec:zero-shot}), leaf-paradigms,text width=5.5em
                        [
                            Zero-shot \\ Prompting, leaf-paradigms, text width=6em
                            [Xie et al.~\cite{xie2024self}{, }QA4RE~\cite{qa4re}{, }Cai et al.~\cite{cai2023monte}{, }AugURE~\cite{augure}{, }Li et al.~\cite{li2023revisiting}{, }Code4Struct~\cite{code4struct}{, }CodeKGC~\cite{codekgc}{, }\\ChatIE~\cite{chatie}{, }SLCoLM~\cite{slcolm}{, }ERA-COT~\cite{era-cot}{, }RAG4RE~\cite{efeoglu2024retrieval}{, }Lin et al.~\cite{LinZS23}{, }Li et al.~\cite{li2024simple}{, }Munnangi et al.~\cite{munnangi2024fly}{, }\\Li et al.~\cite{li2024meta}{, }Li et al.~\cite{li2024empirical}{, }Li et al.~\cite{li2024relation}{, }Sun et al.~\cite{sun2024leveraging}{, }Hu et al.~\cite{hu2024improving}{, }Shao et al.~\cite{shao2024astronomical}, modelnode-paradigms,text width=44.5em]
                        ]
                        [
                            Cross-Domain \\ Learning , leaf-paradigms, text width=6em
                            [DEEPSTRUCT~\cite{deepstruct}{, }X-GEAR~\cite{huang2022multilingual}{, }InstructUIE~\cite{instructuie}{, }UniNER~\cite{universalner}{, }GoLLIE~\cite{gollie}{, }GNER~\cite{gner}{, }KnowCoder~\cite{knowcoder}{, }\\YAYI-UIE~\cite{yayi-uie}{, }ChatUIE~\cite{chatuie}{, }IEPile~\cite{iepile}{, }ULTRA~\cite{zhang2024ultra}{, }Keloth et al.~\cite{keloth2024advancing}{, }Guo et al.~\cite{guo2024diluie}, modelnode-paradigms,text width=44.5em]
                        ]
                        [
                            Cross-Type  Learning , leaf-paradigms,text width=9em
                            [BART-Gen~\cite{bart-gen}, modelnode-paradigms,text width=41.5em]
                        ]
                    ]
                    [
                        Constrained Decoding Generation (\S\ref{sec:decoding}), leaf-paradigms, text width=14.5em
                        [GCD~\cite{geng2023flexible}{, }ASP~\cite{liu2022autoregressive}{, }UIE~\cite{uie}{, }Zaratiana et al.~\cite{ZaratianaTHC24}{, }DORE\cite{dore}, modelnode-paradigms,text width=43em] 
                    ]
                    [
                        Few-shot \\ (\S\ref{sec:fewshot}), leaf-paradigms,text width=5.5em
                        [
                            Fine-tuning , leaf-paradigms, text width=6em
                            [Cui et al.~\cite{cui2021template}{, }TANL~\cite{paolini2021structured}{, }Wang et al.~\cite{instructionner}{, }LightNER~\cite{lightner}{, }UIE~\cite{uie}{, }Cp-NER~\cite{cp-ner}{, }DemoSG~\cite{demosg}{, }KnowCoder~\cite{knowcoder}{, }\\MetaNER~\cite{metaner}{, }Know-Adapter~\cite{nie2024know}{, }Guo et al.~~\cite{guo2024diluie}{, }Hou et al.~~\cite{hou2024knowledge}{, }Duan et al.~~\cite{duan2024keyee}{, }Li et al.~~\cite{li2024meta}, modelnode-paradigms,text width=44.5em]
                        ]
                        [
                            In-Context \\ Learning, leaf-paradigms, text width=6em
                            [GPT-NER~\cite{gpt-ner}{, }Ma et al.~\cite{ma2023large}{, }PromptNER~\cite{promptner}{, }Xie et al.~\cite{xie2024self}{, }QA4RE~\cite{qa4re}{, }GPT-RE~\cite{gpt-re}{, }Xu et al.~\cite{xu2023unleash}{, }\\Code4Struct~\cite{code4struct}{, }CODEIE~\cite{codeie}{, }CodeKGC~\cite{codekgc}{, }GL~\cite{gl}{, }Code4UIE~\cite{code4uie}{, }Cai et al.~\cite{cai2023context}{, }2INER~\cite{2iner}{, }CoT-ER~\cite{cot-er}{, }\\REPLM~\cite{replm}{, }I2CL~\cite{i2cl}{, }LinkNER~\cite{linkner}{, }SLCoLM~\cite{slcolm}{, }C-ICL~\cite{c-icl}{, }ConsistNER~\cite{consistner}{, }RT~\cite{li2024rt}{, }CHisIEC~\cite{chisiec}{, }\\TEXTEE~\cite{Huang23textee}{, }Kwak et al.~\cite{KwakJFBMS23}{, }Bolucu et al.~\cite{bolucu2023impact}{, }Qi et al.~\cite{qi2023mastering}{, }Zhang et al.~\cite{zhang2024context}{, }Naguib et al.~\cite{naguib2024few}{, }\\Munnangi et al.~\cite{munnangi2024fly}{, }Liu et al.~\cite{liu2023improving}{, }Yan et al.~\cite{yan2024ltner}{, }Nunes et al.~\cite{nunesout}{, }Jiang et al.~\cite{jiang2024p}{, }Monajatipoor et al.~\cite{monajatipoor2024llms}{, }\\Otto et al.~\cite{otto2024enhancing}{, }Li et al.~\cite{li2024empirical}{, }Sun et al.~\cite{sun2024leveraging}{, }Zhou et al.~\cite{zhou2023heuristics}{, }Hu et al.~\cite{hu2024improving}{, }Ni et al.~\cite{ni2023unified}, modelnode-paradigms,text width=44.5em]
                        ]
                    ]
                    [
                        Supervised \\ Fine-tuning \\ (\S\ref{sec:full}), leaf-paradigms, text width=5.5em
                        [Yan et al.~\cite{yan2021unified}{, }TEMPGEN~\cite{tempgen}{, }REBEL\cite{rebel}{, }Text2Event~\cite{text2event}{, }Cui et al.~\cite{cui2021template}{, }TANL~\cite{paolini2021structured}{, }ClarET~\cite{claret}{, }DEEPSTRUCT~\cite{deepstruct}{, }\\GTEE-DYNPREF~\cite{gtee-dynpref}{, }GenIE~\cite{genie}{, }PAIE~\cite{paie}{, }UIE~\cite{uie}{, }Xia et al.~\cite{xia2023debiasing}{, }QGA-EE~\cite{qga-ee}{, }InstructUIE~\cite{instructuie}{, }PGAD~\cite{pgad}{, }UniNER~\cite{universalner}{, }\\GoLLIE~\cite{gollie}{, }Set~\cite{set}{, }DemoSG~\cite{demosg}{, }RELA~\cite{rela}{, }DORE~\cite{dore}{, }NAG-NER~\cite{nag-ner}{, }GNER~\cite{gner}{, }KnowCoder~\cite{knowcoder}{, }SPEAE~\cite{speae}{, }AMPERE~\cite{ampere}{, }\\EventRL~\cite{eventrl}{, }YAYI-UIE~\cite{yayi-uie}{, }ChatUIE~\cite{chatuie}{, }PaDeLLM-NER~\cite{padellm-ner}{, }AutoRE~\cite{autore}{, }VANER~\cite{biana2024vaner}{, }CHisIEC~\cite{chisiec}{, }CRE-LLM~\cite{shi2024cre}\\Su et al.~\cite{SuY23}{, }Zaratiana et al.~\cite{ZaratianaTHC24}{, }Fan et al.~\cite{FanH23}{, }Popovivc et al.~\cite{popovivc2024embedded}{, }Keloth et al.~\cite{keloth2024advancing}{, }GLiNER~\cite{gliner}{, }Ding et al.~\cite{ding2024improving}{, }etc.
                        , modelnode-paradigms,text width=52em] 
                    ]
                ]
                [
                    Specific \\ Domains \\ (\S\ref{sec:domain}), leaf-others,text width=5em
                    [ \textbf{\textit{Multimodal}} (Chen et al.~\cite{chen2023chain} PGIM~\cite{pgim} Cai et al.~\cite{cai2023context} RiVEG~\cite{riveg}){, } \textbf{\textit{Multilingual}} (X-GEAR~\cite{huang2022multilingual} Meoni et al.~\cite{meoni2023large} Naguib et al.~\cite{naguib2024few}){, }\textbf{\textit{Scientific}} (Dunn et al.~\cite{dunn2022structured} \\PolyIE~\cite{polyie} Bolucu et al.~\cite{bolucu2023impact} Foppiano et al.~\cite{foppiano2024mining} Dagdelen et al.~\cite{dagdelen2024structured}){, }\textbf{\textit{Medical}} (DICE~\cite{dice} Tang et al.~\cite{tang2023does} Hu et al.~\cite{hu2023zero} Meoni et al.~\cite{meoni2023large} Bian et al.~\cite{bian2023inspire} \\RT~\cite{li2024rt} Agrawal et al.~\cite{agrawal2022large} Labrak et al.~\cite{labrak2023zero} Gutierrez et al.~\cite{gutierrez2022thinking} VANER~\cite{vaner} Munnangi et al.~\cite{munnangi2024fly} Monajatipoor et al.~\cite{monajatipoor2024llms} Hu et al.~\cite{hu2024improving}){, } \\\textbf{\textit{Astronomical}} (Shao et al.~\cite{shao2024astronomical} Evans et al.~\cite{evans2024astro}){, }\textbf{\textit{Historical}} (Gonzalez et al.~\cite{gonzalez2023yes} CHisIEC\cite{chisiec}){, }\textbf{\textit{Legal}} (Nune et al.~\cite{nunesout} Oliveira et al.~\cite{oliveira2024combining} Kwak et al.~\cite{KwakJFBMS23}), modelnode-others, text width=59em]
                ]
                [
                    Evaluation  \\  \& Analysis \\ (\S\ref{sec:eval}), leaf-others,text width=5em
                    [Gutierrez et al.~\cite{gutierrez2022thinking}{, }GPT-3+R~\cite{agrawal2022large}{, }Labrak et al.~\cite{labrak2023zero}{, }Xie et al.~\cite{xie2023empirical}{, }Gao et al.~\cite{gao2023exploring}{, }RT~\cite{li2024rt}{, }InstructIE~\cite{instructie}{, }Han et al.~\cite{han2023information}{, }Katz et al.~\cite{katz2023neretrieve}{, }Hu et al.~\cite{hu2023zero}{, }\\Gonzalez et al.~\cite{gonzalez2023yes}{, }Yuan et al.~\cite{yuan2023zero}{, }Wadhwa et al.~\cite{wadhwa2023revisiting}{, }PolyIE~\cite{polyie}{, }Li et al.~\cite{li2023evaluating}{, }Qi et al.~\cite{qi2023preserving}{, }XNLP~\cite{fei2023xnlp}{, }IEPile~\cite{iepile}{, }Naguib et al.~\cite{naguib2024few}{, }\\Evans et al.~\cite{evans2024astro}{, }Monajatipoor et al.~\cite{monajatipoor2024llms}{, }Atuhurra et al.~\cite{atuhurra2024distilling}{, }CHisIEC~\cite{chisiec}{, }Li et al.~\cite{li2024relation}{, }TEXTEE~\cite{Huang23textee}{, }Foppiano et al.~\cite{foppiano2024mining}, modelnode-others,text width=59em]
                ]
            ]
        \end{forest}
    }
    \caption{Taxonomy of research in generative IE using LLMs. Some papers have been omitted due to space limitations.}
    \label{fig_taxonomy}
    \vspace{-10pt}
\end{figure*}
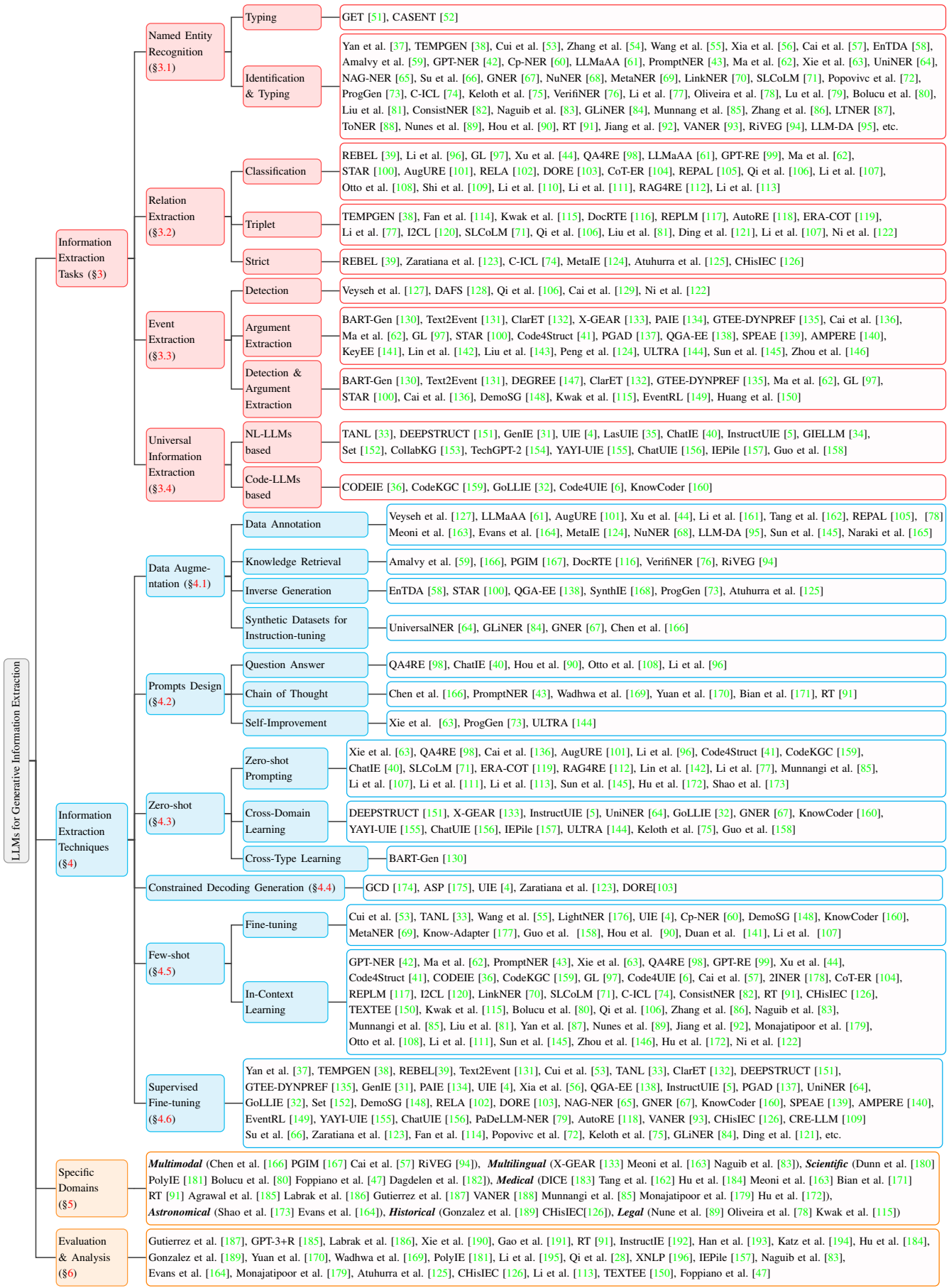

\section{Preliminaries of Generative IE}\label{sec_prelimi}
In this section, we provide a formal definition of discriminative and generative IE and summarize the IE subtasks, as outlined in \cite{ye2022generative}. This survey focuses primarily on the tasks of Named Entity Recognition (NER), Relation Extraction (RE), and Event Extraction (EE) \cite{instructuie,gollie}, as these are tasks that receive the most attention in IE papers. Examples are shown in Fig. \ref{fig:cases}.

(1) For a discriminative model, the objective is to maximize the likelihood of the data. This involves considering an annotated sentence $x$ and a collection of potentially overlapping triples. $t_j = {(s, r, o)}$:
\begin{equation}
p_{cls}(t|x) = \prod_ {(s,r,o) \in t_j} p ((s, r, o) | x_j )
\end{equation}

Another method of discrimination involves generating tags using sequential tagging for each position $i$. For a sentence $x$ consisting of n words, n different tag sequences are annotated based on the "BIESO" (Begin, Inside, End, Single, Outside) notation schema. During the training of the model, the objective is to maximize the log-likelihood of the target tag sequence by utilizing the hidden vector $h_i$  at each position $i$:
\begin{equation}
p_{tag}(y | x) = \frac{exp(h_i,y_i)}{exp(exp(h_i,y'_i))}
\end{equation}

(2) The three types of IE tasks can be formulated in a generative manner. Given an input text (e.g., sentence or document) with a sequence of $n$ tokens $\mathcal{X} = [x_1,...,x_n]$, a prompt $\mathcal{P}$, and the target extraction sequence $\mathcal{Y} = [y_1,...,y_m]$, the objective is to maximize the conditional probability in an auto-regressive formulation:
\begin{equation}
    p_\theta(\mathcal{Y}|\mathcal{X},\mathcal{P}) = \prod_{i=1}^{m}p_\theta(y_i|\mathcal{X},\mathcal{P}, y_{<i}),
\end{equation}
where $\theta$ donates the parameters of LLMs, which can be frozen or trainable. In the era of LLMs, several works have proposed appending extra prompts or instructions $\mathcal{P}$ to $\mathcal{X}$ to enhance the comprehensibility of the task for LLMs \cite{instructuie}.
Even though the input text $\mathcal{X}$ remains the same, the target sequence varies for each task:
\begin{figure}[t]
\begin{center}
\includegraphics[width=\linewidth]{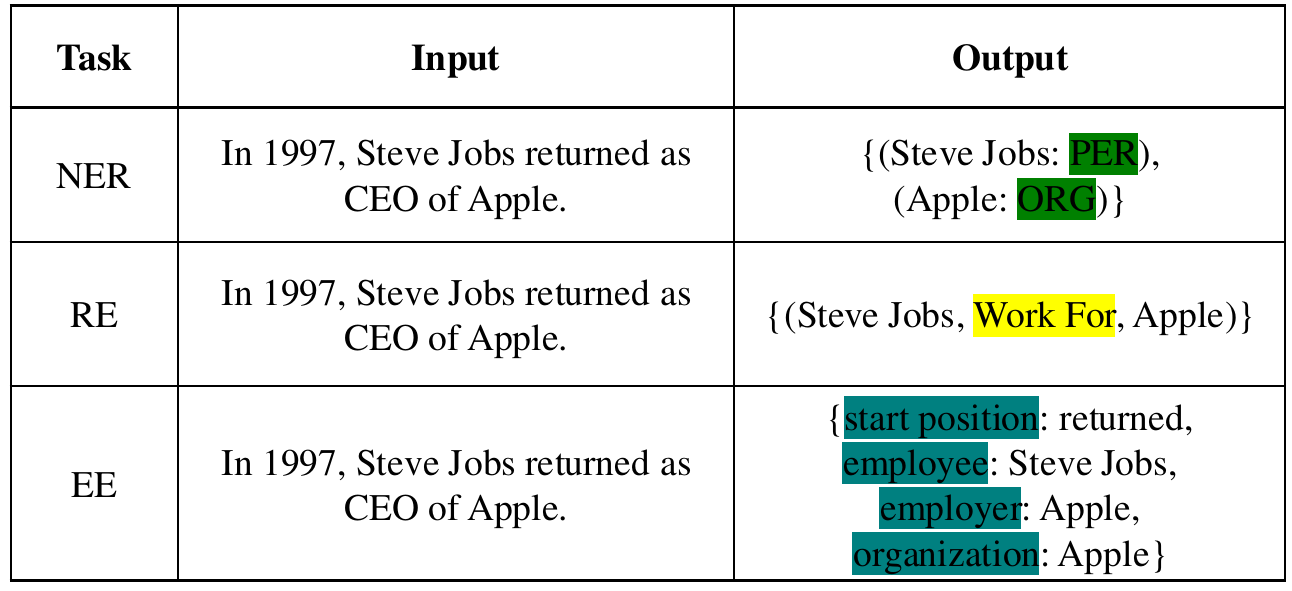} 
\caption{Examples of different IE tasks.}
\label{fig:cases}
\end{center}
\end{figure}
\begin{itemize}
    \item  \textbf{Named Entity Recognition} (NER) includes two tasks: \textbf{Entity Identification} and \textbf{Entity Typing}. The former task is concerned with identifying spans of entities, and the latter task focuses on assigning types to these identified entities.
    \item \textbf{Relation Extraction} (RE) may have different settings in different works. We categorize it using three terms following the literature \cite{uie,instructuie}:
   (1) \textbf{Relation Classification} refers to classifying the relation type between two given entities;  
   (2) \textbf{Relation Triplet} refers to identifying the relation type and the corresponding head and tail entity spans; 
    (3) \textbf{Relation Strict} refers to giving the correct relation type, the span, and the type of head and tail entity.
    \item \textbf{Event Extraction} (EE) can be divided into two subtasks \cite{deepstruct}: 
(1) \textbf{Event Detection} (also known as Event Trigger Extraction in some works) aims to identify and classify the trigger word and type that most clearly represents the occurrence of an event.
(2) \textbf{Event Arguments Extraction} aims to identify and classify arguments with specific roles in the events from the sentences.

\end{itemize}

\section{LLMs for Different Information Extraction Tasks}\label{sec_task}
In this section, we first present a introduction to the relevant LLM technologies for IE subtasks, including NER (\S\ref{sec_ner}), RE (\S\ref{sec_re}), and EE (\S\ref{sec_ee}). We also conduct experimental analysis to evaluate the performance of various methods on representative datasets for three subtasks. Furthermore, we categorize universal IE frameworks into two categories: natural language (NL-LLMs)  and code language (Code-LLMs), to discuss how they model the three distinct tasks using a unified paradigm (\S\ref{sec_uniie}). 

\begin{table*}[t]
\caption{Comparison of LLMs for named entity recognition (identification and typing) with the Micro-F1 metric (\%). $^\dag$ indicates that the model is discriminative. We demonstrate some universal and discriminative models for comparison. IE techniques include Cross-Domain Learning (\textbf{CDL}), Zero-Shot Prompting (\textbf{ZS Pr}), In-Context Learning (\textbf{ICL}), Supervised Fine-Tuning (\textbf{SFT}), Data Augmentation (\textbf{DA}). \textbf{Uni.} denotes whether the model is universal. Onto. 5 denotes the OntoNotes 5.0. Details of datasets and backbones are presented in Section \ref{sec:Benchmarks}. The settings for all subsequent tables are consistent with this format.}
\centering
\setlength\tabcolsep{3pt}  
\begin{tabular}{@{}l|ccl|ccccc@{}}
\toprule
\textbf{Representative Model} &
  \textbf{Paradigm} &
  \textbf{Uni.} &
  \textbf{Backbone} &
  \textbf{ACE04} &
  \textbf{ACE05} &
  \textbf{CoNLL03} &
  \textbf{Onto. 5} & 
  \textbf{GENIA} \\ \midrule
DEEPSTRUCT{\scriptsize ~\cite{deepstruct}}           & CDL    &   & GLM-10B     &   -  & 28.1  & 44.4  & 42.5  & 47.2  \\
Xie et al.{\scriptsize~\cite{xie2024self}}                      & ZS Pr    &  & GPT-3.5-turbo     &   -   & 32.27 & 74.51 &  -   & 52.06 \\
CODEIE{\scriptsize~\cite{codeie}}                    & ICL     & $\surd$ & Code-davinci-002  & 55.29 & 54.82 & 82.32 & -  &  -  \\
Code4UIE{\scriptsize~\cite{code4uie}}                & ICL     & $\surd$ & Text-davinci-003  & 60.1  & 60.9  & 83.6  & -  &  - \\
PromptNER{\scriptsize~\cite{promptner}}              & ICL     &    & GPT-4      &   -   &    -   & 83.48 &   -   & 58.44 \\
Xie et al.{\scriptsize~\cite{xie2024self}}                      & ICL     &    & GPT-3.5-turbo     &  -   & 55.54 & 84.51 &   -    & 58.72 \\
GPT-NER{\scriptsize ~\cite{gpt-ner}}                  & ICL     &   & Text-davinci-003  & 74.2  & 73.59 & 90.91 & 82.2  & 64.42 \\
TANL{\scriptsize ~\cite{paolini2021structured}}       & SFT      & $\surd$ & T5-base     &   -    & 84.9  & 91.7  & 89.8  & 76.4  \\
Cui et al.{\scriptsize~\cite{cui2021template}}                  & SFT      &   & BART              &   -  &  -   & 92.55 &   -    &   -   \\
Yan et al.{\scriptsize~\cite{yan2021unified}}                   & SFT      &   & BART-large      & 86.84 & 84.74 & 93.24 & 90.38 & 79.23 \\
UIE{\scriptsize ~\cite{uie}}                          & SFT      & $\surd$ & T5-large   & 86.89 & 85.78 & 92.99 &  -  &  -    \\
DEEPSTRUCT{\scriptsize ~\cite{deepstruct}}            & SFT      & $\surd$ & GLM-10B    &   -    & 86.9  & 93.0    & 87.8  & 80.8  \\
Xia et al.{\scriptsize~\cite{xia2023debiasing}}                 & SFT      &   & BART-large      & 87.63 & 86.22 & 93.48 & 90.63 & 79.49 \\
InstructUIE{\scriptsize ~\cite{instructie}}           & SFT      & $\surd$ & Flan-T5-11B  &   -  & 86.66 & 92.94 & 90.19 & 74.71 \\
UniNER{\scriptsize ~\cite{universalner}}              & SFT      &   & LLaMA-7B        & 87.5  & 87.6  &   -    & 89.1  & 80.6  \\
GoLLIE{\scriptsize ~\cite{gollie}}                    & SFT      & $\surd$ & Code-LLaMA-34B  &  -  & 89.6  & 93.1  & 84.6  &  -  \\
EnTDA{\scriptsize ~\cite{entda}}                      & DA       &         & T5-base   & 88.21 & 87.56 & 93.88 & 91.34 & 82.25 \\ 
YAYI-UIE{\scriptsize ~\cite{yayi-uie}} & SFT & $\surd$ & Baichuan2-13B & - & 81.78 & 96.77 & 87.04 & 75.21 \\
ToNER{\scriptsize ~\cite{jiang2024toner}} & SFT & & Flan-T5-3B & 88.09 & 86.68 & 93.59 & 91.30 & - \\
KnowCoder{\scriptsize ~\cite{knowcoder}} & SFT & $\surd$ & LLaMA2-7B & 86.2	& 86.1	& 95.1	& 88.2 & 76.7 \\
GNER{\scriptsize ~\cite{gner}} & SFT & & Flan-T5-11B & -	& -	& 93.28	& 91.83	& - \\
\midrule
USM$^\dag${\scriptsize ~\cite{usm}}                 & SFT   & $\surd$ & RoBERTa-large     & 87.62 & 87.14 & 93.16 &  -  & -  \\
RexUIE$^\dag${\scriptsize ~\cite{rexuie}}           & SFT   & $\surd$ & DeBERTa-v3-large  & 87.25 & 87.23 & 93.67 &  -  & -  \\
Mirror$^\dag${\scriptsize ~\cite{mirror}}           & SFT   & $\surd$ & DeBERTa-v3-large  & 87.16 & 85.34 & 92.73 &  -  & -  \\   
\bottomrule
\end{tabular}
\label{tab:ner-table}
\vspace{-10pt}
\end{table*}

\subsection{Named Entity Recognition}\label{sec_ner}
NER is a crucial component of IE and can be seen as a predecessor or subtask of RE and EE. It is also a fundamental task in other NLP tasks, thus attracting significant attention from researchers to explore new possibilities in the era of LLMs~\cite{hou2024knowledge,li2024rt,jiang2024p,evans2024astro,biana2024vaner,monajatipoor2024llms,otto2024enhancing,naraki2024augmenting,peng2024metaie,atuhurra2024distilling,chisiec,ye2024llm,riveg,hu2024improving,ni2023unified,nie2024know,foppiano2024mining,dagdelen2024structured,shao2024astronomical}.
Considering the gap between the sequence labeling and generation models, GPT-NER \cite{gpt-ner} transformed NER into a generative task and proposed a self-verification strategy to rectify the mislabeling of NULL inputs as entities.
Xie et al.~\cite{xie2024self} proposed a training-free self-improving framework that uses LLM to predict on the unlabeled corpus to obtain pseudo demonstrations, thereby enhancing the performance of LLM on zero-shot NER.

Table \ref{tab:ner-table} shows the comparison of NER on five main datasets, which are obtained from their original papers. 
We can observe that: 
\begin{itemize}
    \item 1) the models in few-shot and zero-shot settings still have a huge performance gap behind the SFT and DA.
    \item 2) Even though there is little difference between backbones, the performance varies greatly between methods under the ICL paradigm. For example, GPT-NER opens up at least a 6\% F1 value gap with other methods on each dataset, and up to about 19\% higher.
    \item 3) Compared to ICL, there are only minor differences in performance between different models after SFT, even though the parameters in their backbones can differ by up to a few hundred times.
    \item 4) The performance of models trained with the SFT paradigm exhibits greater variability across datasets, particularly for universal models. For instance, YAYI-UIE~\cite{yayi-uie} and KnowCoder~\cite{knowcoder} outperform other models by at least 2.89\% and 1.22\% respectively on CoNLL03, while experiencing a decrease of 7.04\% and 5.55\% respectively compared to the best model on GENIA. We hypothesize that this discrepancy may arise from these models being trained on diverse datasets primarily sourced from news and social media domains, whereas GENIA represents a smaller fraction in the training set as it belongs to the biomedical domain; thus resulting in significant distribution gaps between different fields that ultimately impact performance outcomes. Furthermore, universal models necessitate simultaneous training across multiple subtasks, which inevitably exacerbates this distribution gap.
    \item 5) The EnTDA~\cite{entda}, on the contrary, exhibits exceptional stability and outperforms other methods on all datasets, thereby substantiating the robustness of the DA paradigm in addressing specific tasks.
\end{itemize}

\begin{table*}[t]
\caption{Comparison of LLMs for relation extraction with the ``relation strict'' \cite{uie} Micro-F1 metric (\%). $^\dag$ indicates that the model is discriminative.}
\centering
\setlength\tabcolsep{4pt}  
{ 
\begin{tabular}{@{}l|ccl|ccccc@{}}
\toprule
\textbf{Representative Model} &
  \textbf{Technique} &
  \textbf{Uni.} &
  \textbf{Backbone} &
  \textbf{NYT} &
  \textbf{ACE05} &
  \textbf{ADE} &
  \textbf{CoNLL04} &
  \textbf{SciERC} \\ 
\midrule
CodeKGC{\scriptsize ~\cite{codekgc}}         & ZS Pr   & $\surd$ & Text-davinci-003 &   -    &    -   & 42.8  & 35.9  & 15.3  \\
CODEIE{\scriptsize ~\cite{codeie}}           & ICL   & $\surd$ & Code-davinci-002 & 32.17 & 14.02 &    -   & 53.1  & 7.74  \\
CodeKGC{\scriptsize ~\cite{codekgc}}         & ICL   & $\surd$ & Text-davinci-003 &   -    &    -   & 64.6  & 49.8  & 24.0  \\
Code4UIE{\scriptsize ~\cite{code4uie}}       & ICL   & $\surd$ & Text-davinci-002 & 54.4  & 17.5  & 58.6  & 54.4  &   -    \\
REBEL{\scriptsize ~\cite{rebel}}             & SFT   &       & BART-large       & 91.96 &   -    & 82.21 & 75.35 &   -    \\
UIE{\scriptsize ~\cite{uie}}                 & SFT & $\surd$ & T5-large         &    -   & 66.06 &   -    & 75.0  & 36.53 \\
InstructUIE{\scriptsize ~\cite{instructuie}} & SFT & $\surd$ & Flan-T5-11B      & 90.47 &    -   & 82.31 & 78.48 & 45.15 \\ 
GoLLIE{\scriptsize ~\cite{gollie}}           & SFT & $\surd$ & Code-LLaMA-34B   &  -     & 70.1  &    -   &    -   &   -    \\ 
YAYI-UIE{\scriptsize \cite{yayi-uie}} & SFT & $\surd$ & Baichuan2-13B & 89.97	& -	& 84.41	& 79.73	& 40.94 \\
KnowCoder{\scriptsize ~\cite{knowcoder}} & SFT & $\surd$ & LLaMA2-7B & 93.7	& 64.5	& 84.8	& 73.3 & 40.0 \\
\midrule
USM$^\dag${\scriptsize ~\cite{usm}}          & SFT & $\surd$ & RoBERTa-large    & - & 67.88 &   -    & 78.84 & 37.36 \\
RexUIE$^\dag${\scriptsize ~\cite{rexuie}}    & SFT & $\surd$ & DeBERTa-v3-large & - & 64.87 &   -    & 78.39 & 38.37 \\
\bottomrule
\end{tabular}
}
\label{tab:re-table}
\vspace{-8pt}
\end{table*}


\begin{table*}[t]
\caption{Comparison of LLMs for relation classification with the Micro-F1 metric (\%). }
\centering
\setlength\tabcolsep{3pt}  
\begin{tabular}{@{}l|ccl|cccc@{}}
\toprule
\textbf{Representative Model} &
  \textbf{Technique} &
  \textbf{Uni.} &
  \textbf{Backbone} &
  \textbf{TACRED} &
  \textbf{Re-TACRED} &
  \textbf{TACREV} &
  \textbf{SemEval}  \\ \midrule
QA4RE{\scriptsize~\cite{qa4re}}         & ZS Pr &  & Text-davinci-003 & 59.4 & 61.2  & 59.4  & 43.3 \\
SUMASK{\scriptsize~\cite{li2023revisiting}}   & ZS Pr &  & GPT-3.5-turbo-0301  & 79.6 &  73.8 & 75.1 & - \\
GPT-RE{\scriptsize~\cite{gpt-re}}       & ICL &  & Text-davinci-003 & 72.15 & -  &  -   & 91.9  \\
Xu et al.{\scriptsize~\cite{xu2023unleash}}      & ICL &  & Text-davinci-003 & 31.0  & 51.8 & 31.9 & -\\
REBEL{\scriptsize~\cite{rebel}}         & SFT   &  & BART-large   & - & 90.36 & - & - \\ 
Xu et al.{\scriptsize~\cite{xu2023unleash}}      & DA    &  & Text-davinci-003 & 37.4 & 66.2 & 41.0  & -\\
\bottomrule
\end{tabular}
\vspace{-15pt}
\label{tab:rc-table}
\end{table*}

\subsection{Relation Extraction}\label{sec_re}
RE also plays an important role in IE, which usually has different setups in different studies as mentioned in Section \ref{sec_prelimi}. 
To address the poor performance of LLMs on RE tasks due to the low incidence of RE in instruction-tuning datasets, as indicated by \cite{gutierrez2022thinking}, QA4RE~\cite{qa4re} introduced a framework that enhances LLMs' performance by aligning RE tasks with QA tasks.
GPT-RE~\cite{gpt-re} incorporates task-aware representations and enriching demonstrations with reasoning logic to improve the low relevance between entity and relation and the inability to explain input-label mappings.
Due to the large number of predefined relation types and uncontrolled LLMs, Li et al.~\cite{li2023semi} proposed to integrate LLM with a natural language inference module to generate relation triples, enhancing document-level relation datasets.

As shown in the Table \ref{tab:re-table} and \ref{tab:rc-table}, we statistically found that universal IE models are generally better solving harder Relation Strict problems due to learning the dependencies between multi-tasks \cite{paolini2021structured, uie}, while the task-specific methods solve simpler RE subtasks (e.g. relation classification).
Moreover, when compared to NER, it becomes evident that the performance disparities among models in RE are more pronounced, thereby highlighting the potential of LLM in addressing RE tasks.

\begin{table*}[t]
\caption{Comparison of Micro-F1 Values for Event Extraction on ACE05. Evaluation tasks include: Trigger Identification (Trg-I), Trigger Classification (Trg-C),  Argument Identification (Arg-I), and Argument Classification (Arg-C). $^\dag$ indicates that the model is discriminative.}
\centering
\setlength\tabcolsep{8pt}  
\begin{tabular}{@{}l|ccl|ccccc@{}}
\toprule
\textbf{Representative Model} & \textbf{Technique} & \textbf{Uni.} & \textbf{Backbone} & \textbf{Trg-I} & \textbf{Trg-C} & \textbf{Arg-I} & \textbf{Arg-C} \\ \midrule
Code4Struct{\scriptsize ~\cite{code4struct}}  & ZS Pr   &      & Code-davinci-002  &   -    &    -  & 50.6 & 36.0  \\
Code4UIE{\scriptsize ~\cite{code4uie}}     & ICL   & $\surd$ & GPT-3.5-turbo-16k &   -    & 37.4  &   -   & 21.3  \\
Code4Struct{\scriptsize ~\cite{code4struct}}  & ICL   &      & Code-davinci-002  &   -    &   -    & 62.1 & 58.5  \\
TANL{\scriptsize ~\cite{paolini2021structured}}         & SFT & $\surd$ & T5-base    & 72.9  & 68.4  & 50.1 & 47.6  \\
Text2Event{\scriptsize ~\cite{text2event}}   & SFT &     & T5-large          &   -    & 71.9  &   -   & 53.8  \\
BART-Gen{\scriptsize ~\cite{bart-gen}}     & SFT &     & BART-large        &    -   &   -    & 69.9 & 66.7  \\
UIE{\scriptsize ~\cite{uie}}          & SFT & $\surd$ & T5-large          &    -   & 73.36 &   -   & 54.79 \\
GTEE-DYNPREF{\scriptsize ~\cite{gtee-dynpref}} & SFT &       & BART-large        &   -    & 72.6  &   -   & 55.8  \\ 
DEEPSTRUCT{\scriptsize ~\cite{deepstruct}}   & SFT & $\surd$ & GLM-10B           & 73.5  & 69.8  & 59.4 & 56.2  \\
PAIE{\scriptsize ~\cite{paie}}         & SFT &       & BART-large        &   -    &   -    & 75.7 & 72.7  \\
PGAD{\scriptsize ~\cite{pgad}}         & SFT &      & BART-base         &    -   &    -   & 74.1 & 70.5  \\
QGA-EE{\scriptsize ~\cite{qga-ee}}       & SFT &         & T5-large          &   -    &   -    & 75.0 & 72.8  \\
InstructUIE{\scriptsize ~\cite{instructuie}}  & SFT & $\surd$ & Flan-T5-11B       &    -   & 77.13 &   -   & 72.94 \\
GoLLIE{\scriptsize ~\cite{gollie}}       & SFT & $\surd$ & Code-LLaMA-34B    &    -   & 71.9  &   -   & 68.6  \\
YAYI-UIE{\scriptsize ~\cite{yayi-uie}} & SFT & $\surd$ & Baichuan2-13B & - & 65.0	& - &62.71 \\
KnowCoder{\scriptsize ~\cite{knowcoder}} & SFT & $\surd$ & LLaMA2-7B & - & 74.2	& - & 70.3 \\
\midrule
USM$^\dag${\scriptsize ~\cite{usm}}           & SFT & $\surd$ & RoBERTa-large     &    -   & 72.41 &    -  & 55.83 \\
RexUIE$^\dag${\scriptsize ~\cite{rexuie}}        & SFT & $\surd$ & DeBERTa-v3-large  &   -    & 75.17 &   -   & 59.15 \\
Mirror$^\dag${\scriptsize ~\cite{mirror}}       & SFT & $\surd$ & DeBERTa-v3-large  &    -   & 74.44 &    -  & 55.88 \\ 
\bottomrule
\end{tabular}
\label{tab:ee-table}
\vspace{-10pt}
\end{table*}

\subsection{Event Extraction}\label{sec_ee} Events can be defined as specific occurrences or incidents that happen in a given context. Recently, many studies \cite{gtee-dynpref, qga-ee} aim to understand events and capture their correlations by extracting event triggers and arguments using LLMs, which is essential for various reasoning tasks \cite{Bhagavatula2020Abductive}.
For example, Code4Struct~\cite{code4struct} leveraged LLMs to translate text into code to tackle structured prediction tasks, using programming language features to introduce external knowledge and constraints through alignment between structure and code.
Considering the interrelation between different arguments in the extended context, 
PGAD~\cite{pgad} employed a text diffusion model to create a variety of context-aware prompt representations, enhancing both sentence-level and document-level event argument extraction by identifying multiple role-specific argument span queries and coordinating them with context.

As can be seen from results of recent studies in Table \ref{tab:ee-table}, vast majority of current methods are based on SFT paradigm, and only a few methods that use LLMs for either zero-shot or few-shot learning.
In addition, generative methods outperform discriminative ones by a wide margin, especially in metric for argument classification task, indicating the great potential of generative LLMs for EE.

\begin{figure*}[h]
\begin{center}
\includegraphics[scale=0.75]{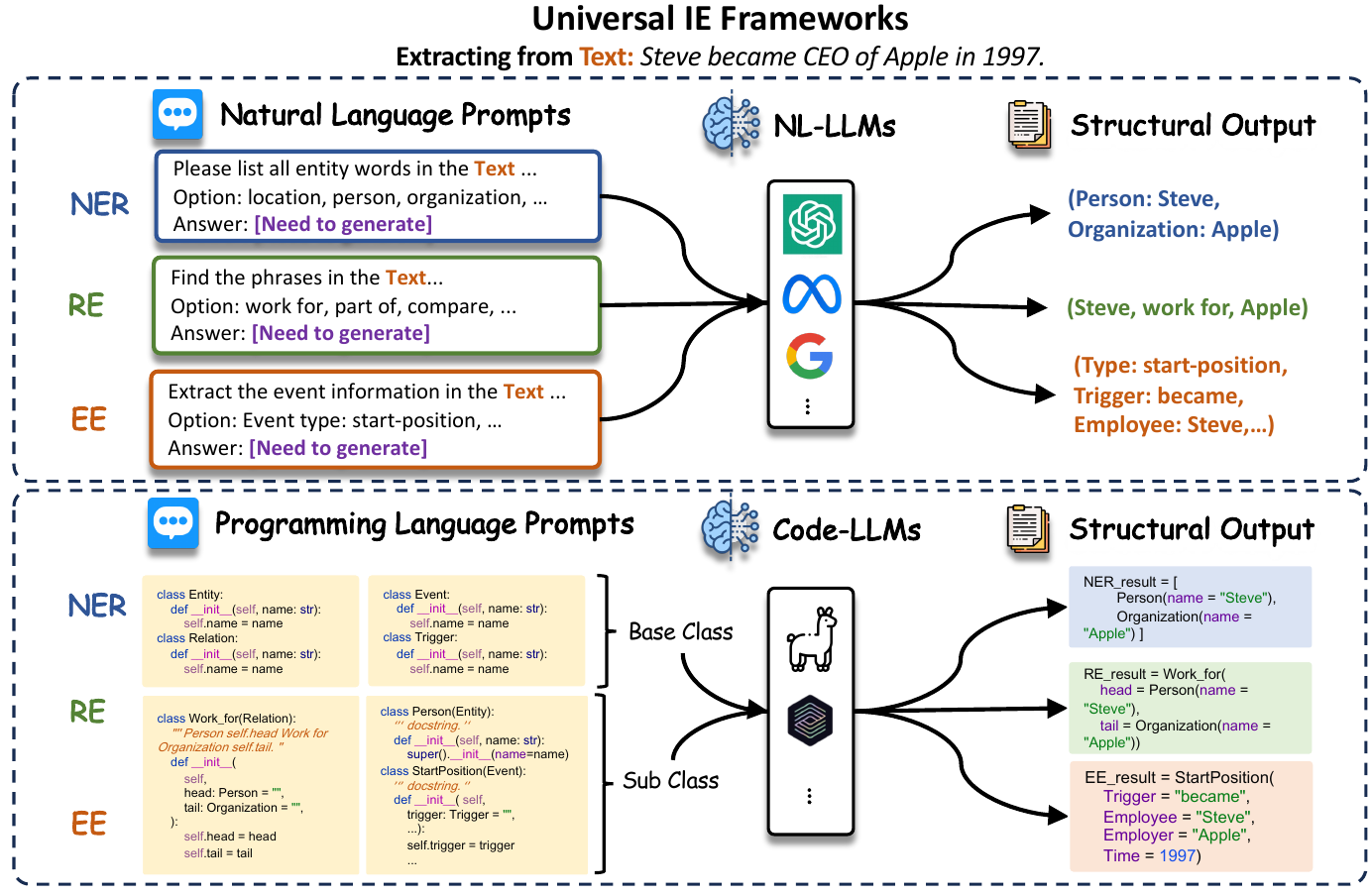} 
\caption{The comparison of prompts of NL-LLMs and Code-LLMs for universal IE. Both NL-based and code-based methods attempt to construct a universal schema, but they differ in terms of prompt format and the way they utilize the generation capabilities of LLMs. This figure is adopted from \cite{instructuie} and \cite{code4uie}.}
\label{fig:tasks}
\end{center}
\vspace{-15pt}
\end{figure*}

\subsection{Universal Information Extraction}\label{sec_uniie}
Different IE tasks vary a lot, with different optimization objectives and task-specific schemas, requiring separate models to handle the complexity of different IE tasks, settings, and scenarios \cite{uie}. As shown in Fig. \ref{fig_taxonomy}, many works solely focus on a subtask of IE. However, recent advancements in LLMs have led to the proposal of a unified generative framework in several studies \cite{instructuie,gollie}. This framework aims to model all IE tasks, capturing the common abilities of IE and learning the dependencies across multiple tasks. The prompt format for Uni-IE can typically be divided into natural language-based LLMs (NL-LLMs) and code-based LLMs (code-LLMs), as illustrated in Fig. \ref{fig:tasks}.

\noindent
\textbf{NL-LLMs.} NL-based methods unify all IE tasks in a universal natural language schema. For instance, UIE \cite{uie} proposed a unified text-to-structure generation framework that encodes extraction structures, and captured common IE abilities through a structured extraction language. InstructUIE \cite{instructuie} enhanced UIE by constructing expert-written instructions for fine-tuning LLMs to consistently model different IE tasks and capture the inter-task dependency.
Additionally, ChatIE \cite{chatie} explored the use of LLMs like ChatGPT~\cite{chatgpt} in zero-shot prompting, transforming the task into a multi-turn question-answering problem.

\noindent
\textbf{Code-LLMs.} On the other hand, code-based methods unify IE tasks by generating code with a universal programming schema \cite{code4struct}. 
Code4UIE \cite{code4uie} proposed a universal retrieval-augmented code generation framework, which leverages Python classes to define schemas and uses in-context learning to generate codes that extract structural knowledge from texts.
Besides, CodeKGC \cite{codekgc} leveraged the structural knowledge inherent in code and employed the schema-aware prompts and rationale-enhanced generation to improve performance.
To enable LLMs to adhere to guidelines out-of-the-box, GoLLIE \cite{gollie} enhanced zero-shot ability on unseen IE tasks by aligning with annotation guidelines.

In general, NL-LLMs are trained on a wide range of text and can understand and generate human language, which allows the prompts and instructions to be conciser and easier to design. However, NL-LLMs may produce unnatural outputs due to the distinct syntax and structure of IE tasks \cite{codekgc}, which differ from the training data. Code, being a formalized language, possesses the inherent capability to accurately represent knowledge across diverse schemas, which makes it more suitable for structural prediction \cite{code4uie}. But code-based methods often require a substantial amount of text to define a Python class (see Fig. \ref{fig:tasks}), which in turn limits the sample size of the context.
Through experimental comparison in Table \ref{tab:ner-table}, \ref{tab:re-table}, and \ref{tab:ee-table}, we can observe that Uni-IE models after SFT outperform task-specific models in the NER, RE, and EE tasks for most datasets.

\subsection{Summaries of Tasks}

In this section, we explored the three primary tasks within IE and their associated sub-tasks, as well as frameworks that unify these tasks~\cite{uie}. A key observation is the increased application of generative LLMs to NER~\cite{gner,2iner}, which has seen significant advancements and remains a highly active area of research within IE. In contrast, tasks such as relation extraction and event extraction have seen relatively less application, particularly for strict relation extraction~\cite{rebel} and detection-only event extraction~\cite{dafs}. This discrepancy may be attributed to the critical importance of NER, its utility in various downstream tasks, and its relatively simpler structured outputs, which facilitate large-scale supervised fine-tuning~\cite{zhong2023comprehensive}.

Additionally, a notable trend is the emergence of unified models for IE tasks, leveraging the general text understanding capabilities of modern large models~\cite{uie,chatuie,code4uie}. Several studies have proposed unified generative frameworks that capture the common abilities across IE tasks and learn the dependencies between them. These unified approaches can be broadly categorized into natural language-based and code-based methods, each with distinct advantages and limitations.

Experimental results summarized in Tables \ref{tab:ner-table}, \ref{tab:re-table}, \ref{tab:rc-table} and \ref{tab:ee-table} reveal that universal IE models generally perform better on more complex strict relation extraction tasks due to their ability to learn dependencies across multiple tasks. Furthermore, generative methods significantly outperform discriminative ones in event extraction tasks, particularly in argument classification, highlighting the substantial potential of generative LLMs in this domain.

\begin{figure*}[t]
\begin{center}
\includegraphics[scale=0.9]{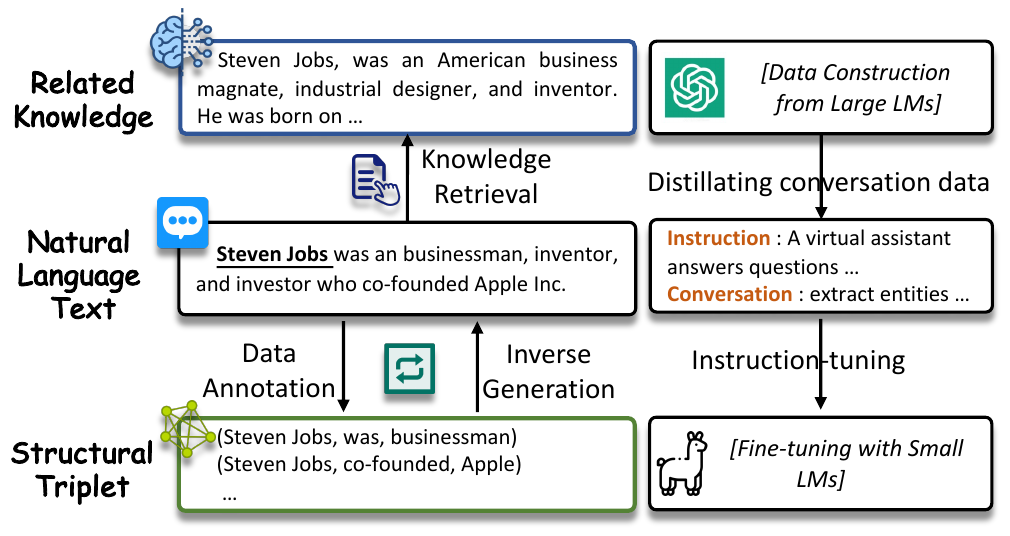} 
\caption{Comparison of data augmentation methods.}
\label{fig:dataaug}
\end{center}
\end{figure*}

\section{Techniques of LLMs for Generative IE}\label{sec:paradigm}
In this section, we categorize recent methods based on their techniques, including \textbf{Data Augmentation} (\S\ref{sec:da}, refers to enhancing information by applying various transformations to the existing data using LLMs), \textbf{Prompt Design} (\S\ref{sec:promptdesign}, refers to the use of task-specific instructions or prompts, to direct the behavior of a model.), \textbf{Zero-shot Learning} (\S\ref{sec:zero-shot}, refers to generating answer without any training examples for the specific IE tasks), \textbf{Constrained Decoding Generation} (\S\ref{sec:decoding}, refers to the process of generating text while adhering to specific constraints or rule), \textbf{Few-shot Learning} (\S\ref{sec:fewshot}, refers to the generalization from a small number of labeled examples by training or in-context learning), \textbf{Supervised Fine-tuning} (\S\ref{sec:full}, refers to further training LLMs on IE tasks using labeled data), to highlight the commonly used approaches for adapting LLMs to IE.

\subsection{Data Augmentation} \label{sec:da}
Data augmentation involves generating meaningful and diverse data to effectively enhance the training examples, while avoiding the introduction of unrealistic, misleading, and offset patterns.
Recent powerful LLMs also demonstrate remarkable performance in data generation tasks \cite{whitehouse2023llm, wu2024survey,chen2024tomgpt,luo2024bridging,yang2024latex}, which has attracted the attention of many researchers using LLMs to generate synthetic data for IE \cite{veyseh2021unleash,llmaaa,augure,xu2023unleash,li2023semi,tang2023does,meoni2023large}.
It can be roughly divided into four strategies according to their techniques, as shown in Fig. \ref{fig:dataaug}. 

\noindent
\textbf{Data Annotation.} This strategy directly generates labeled structural data using LLMs. For instance, Zhang et al.~\cite{llmaaa} proposed LLMaAA to improve accuracy and data efficiency by employing LLMs as annotators within an active learning loop, thereby optimizing both the annotation and training processes. AugURE~\cite{augure} employed within-sentence pairs augmentation and cross-sentence pairs extraction to enhance the diversity of positive pairs for unsupervised RE, and introduced margin loss for sentence pairs. Li et al.~\cite{li2023semi} addressed the challenge of document-level relation extraction from a long context, and proposed an automated annotation method for DocRE that combines a LLM with a natural language inference module to generate relation triples.

\noindent
\textbf{Knowledge Retrieval.} This strategy effectively retrieves related information from LLMs for IE, which is similar to retrieval augmentation generation (RAG) \cite{gao2023retrieval}.
PGIM \cite{pgim} presented a two-stage framework for multimodal NER, which leveraged ChatGPT as an implicit knowledge base to heuristically retrieve auxiliary knowledge for more efficient entity prediction. 
Amalvy et al.~\cite{amalvy2023learning} proposed to improve NER on long documents by generating a synthetic context retrieval training dataset, and training a neural context retriever. 
Chen et al.~\cite{chen2023chain} focused on the task of multimodal NER and RE, and  showcased their approach to enhancing commonsense reasoning skills by employing a range of CoT prompts that encompass different aspects, including nouns, sentences, and multimodal inputs. Additionally, they employed data augmentation techniques such as style, entity, and image manipulation to further improve the performance.

\noindent
\textbf{Inverse Generation.} This strategy encourages LLMs to generate natural text or questions by utilizing the structural data provided as input, which aligns with the training paradigm of LLMs.
For example, SynthIE \cite{synthie} showed that LLMs can create high-quality synthetic data for complex tasks by reversing the task direction, and train new models that outperformed previous benchmarks.
Rather than relying on ground-truth targets, which limits generalizability and scalability, STAR \cite{star} generated structures from valid triggers and arguments, then generates passages with LLMs by designing fine-grained instructions, error identification, and iterative revision. In order to address the challenge of maintaining text coherence while preserving entities, EnTDA \cite{entda} proposed a method that involves manipulating the entity list of the original text. This manipulation includes adding, deleting, replacing, or swapping entities. And it further introduced a diversity beam search to enhance diversity in the Entity-to-text generation process.

\noindent
\textbf{Synthetic Datasets for Fine-tuning.}
This strategy involves generating some data for instruction-tuning by querying LLMs. Typically, this data is generated by a more powerful model for fine-tuning instructions in dialogues, and then distilled onto a smaller model, enabling it to also acquire stronger zero-shot capabilities \cite{universalner,gliner,gner}. For instance, UniversalNER \cite{universalner} explored targeted distillation with mission-focused instruction tuning to train student models that excel in open NER, which used ChatGPT as the teacher model and distilled it into smaller UniversalNER model. GNER \cite{gner} proposed the integration of negative instances to enhance existing methods by introducing contextual information and improving label boundaries. The authors trained their model using Pile-NER, a dataset that includes approximately 240K entities across 13K distinct entity categories, which are sampled from the Pile Corpus \cite{gao2020pile} and processed using ChatGPT to generate the entities. The results demonstrate improved zero-shot performance across unseen entity domains.

\subsection{Prompt Design} \label{sec:promptdesign}
Prompt engineering is a technique employed to enhance the capabilities of LLMs without altering their network parameters \cite{sahoo2024systematic,marvin2023prompt,zhao2024lane,zheng2023generative,wu2024exploring,zheng2024harnessing}. It entails utilizing task-specific instructions, known as prompts, to guide the behavior of model \cite{liu2023pre,chen2023unleashing,zhao2024dynllm}. The practice of prompt design has proven successful in various applications \cite{wang2023prompt,xu2024multi,li2024visualization,liu2024speak}. Undoubtedly, effective prompt design also plays a crucial role in improving the performance of LLMs on IE tasks. In this section, we categorize prompt design approaches based on different strategies and provide a detailed explanation of the underlying motivations behind these techniques:

\noindent
\textbf{Question Answer (QA). }
LLMs are instruction-tuned using a dialogue-based method \cite{peng2023gpt,peng2023you}, which creates a gap when compared to the structured prediction requirements of the IE task. Consequently, recent efforts have been made to employ a QA prompt approach to enhance LLMs and facilitate the generation of desired results more seamlessly \cite{qa4re,chatie,hou2024knowledge,otto2024enhancing,li2023revisiting}.
For example, 
QA4RE \cite{qa4re} found that LLMs tend to perform poorly on RE because the instruction-tuning datasets used to train them have a low incidence of RE tasks, and thus proposes reformulating RE as multiple-choice QA to take advantage of the higher prevalence of QA tasks in instruction-tuning datasets.
Li et al.~\cite{li2023revisiting} analyzed the limitations of existing RE prompts and proposed a new approach called summarize-and-ask prompting, which transforms zero-shot RE inputs into effective QA format using LLMs recursively. It also showed promise in extracting overlapping relations and effectively handling the challenge of none-of-the-above relations.
ChatIE \cite{chatie} proposed a two-stage framework to transform the zero-shot IE task into a multi-turn QA problem. The framework initially identified the different types of elements, and then a sequential IE process is executed for each identified element type. Each stage utilized a multi-turn QA process, where prompts are constructed using templates and previously extracted information.

\noindent
\textbf{Chain-of-thought (CoT). }
CoT \cite{wei2022chain} is a prompting strategy used with LLMs to enhance their performance, by providing a step-wise and coherent reasoning chain as a prompt to guide the model's response generation. CoT prompting has gained attention in recent years \cite{chu2023survey}, and there is ongoing research exploring its effectiveness on IE tasks \cite{chen2023chain,promptner,wadhwa2023revisiting,yuan2023zero,bian2023inspire,li2024rt}. 
PromptNER \cite{promptner} combined LLMs with prompt-based heuristics and entity definitions. It prompted an LLM to generate a list of potential entities and their explanations based on provided entity type definitions.
Bian et al.~\cite{bian2023inspire} proposed a two-step approach to improve Biomedical NER using LLMs. Their approach involved leveraging CoT to enable the LLM to tackle the Biomedical NER task in a step-by-step manner, breaking it down into entity span extraction and entity type determination.
Yuan et al.~\cite{yuan2023zero} also proposed CoT prompt as a two-stage approach to guide ChatGPT in performing temporal relation reasoning for temporal RE task.

\noindent
\textbf{Self-Improvement. }
While COT technology can partially elicit the reasoning ability of LLM, it is unavoidable that LLM will still generate factual errors. As a result, there have been efforts \cite{xie2024self,proggen,zhang2024ultra} to employ LLMs for iterative self-verification and self-improvement, aiming to rectify the results. For instance, Xie et al.~\cite{xie2024self} proposed a training-free self-improving framework, which consists of three main steps. First, LLMs made predictions on unlabeled corpus, generating self-annotated dataset through self-consistency. Second, the authors explored different strategies to select reliable annotations. Finally, during inference, demonstrations from reliable self-annotated dataset were retrieved for in-context learning.
ProgGen \cite{proggen} involved guiding LLMs to engage in self-reflection within specific domains, resulting in the generation of domain-relevant attributes that contribute to the creation of training data enriched with attributes. Additionally, ProgGen employd a proactive strategy by generating entity terms in advance and constructing NER context data around these entities, thereby circumventing the challenges LLMs face when dealing with intricate structures.

\subsection{Zero-shot Learning} \label{sec:zero-shot}
The primary challenges in zero-shot learning involve ensuring that the model can effectively generalize to tasks and domains it has not been trained on, while also aligning the pre-training paradigm of LLMs to these novel tasks.
Due to the large amount of knowledge embedded within, LLMs show impressive abilities in zero-shot scenarios of unseen tasks \cite{kojima2022large, chatie}.
To achieve zero-shot cross-domain generalization of LLMs in IE tasks, several works have been proposed \cite{gollie,universalner,instructuie}. 
These works offered a universal framework for modeling various IE tasks and domains, and introduced innovative training prompts, e.g., instruction \cite{instructuie} and guidelines \cite{gollie}, for learning and capturing the inter-task dependencies of known tasks and generalizing them to unseen tasks and domains.
%
For cross-type generalization, BART-Gen \cite{bart-gen} introduced a document-level neural model that frames the EE task as conditional generation, which led to improved performance and strong portability to unseen event types.

On the other hand, in order to improve the ability of LLMs under zero shot prompts (no need for fine-tuning), QA4RE \cite{qa4re} and ChatIE \cite{chatie} proposed to transform IE into a multi-turn question-answering problem for aligning it with QA task, which is a predominant task
in instruction-tuning datasets.
Li et al.~\cite{li2023revisiting} integrated the chain-of-thought approach and proposed the summarize-and-ask prompting to solve the challenge of ensuring the reliability of outputs from black box LLMs \cite{ma2023large}.

\subsection{Constrained Decoding Generation} \label{sec:decoding}
LLMs are pretrained models that are initially trained on the task of predicting the next token in a sequence. This pretraining allows researchers to leverage the advantages of these models for various NLP tasks \cite{yin2023survey,zhou2022survey}. However, LLMs are primarily designed for generating free-form text and may not perform well on structured prediction tasks where only a limited set of outputs are valid.

To address this challenge, researchers have explored the use of constrained generation for better decoding  \cite{geng2023flexible,liu2022autoregressive,uie,ZaratianaTHC24}. Constrained decoding generation in autoregressive LLMs refers to the process of generating text while adhering to specific constraints or rules \cite{willard2023efficient,beurer2023prompt,zheng2023efficiently}. 
For example, Geng et al.~\cite{geng2023flexible} proposed using grammar-constrained decoding as a solution to control the generation of LMs, ensuring that the output follows a given structure. The authors introduced input-dependent grammars to enhance flexibility, allowing the grammar to depend on the input and generate different output structures for different inputs. 
Unlike previous methods, which generate information token by token, Zaratiana et al.~\cite{ZaratianaTHC24} introduced a new approach for extracting entities and relations by generating a linearized graph with nodes representing text spans and edges representing relation triplets. They used a transformer encoder-decoder architecture with a pointing mechanism and a dynamic vocabulary of spans and relation types, to capture the structural characteristics and boundaries while grounding the output in the original text.

\subsection{Few-shot Learning}\label{sec:fewshot}
Few-shot learning has access to only a limited number of labeled examples, leading to challenges like overfitting and difficulty in capturing complex relationships \cite{huang2020few}.
Fortunately, scaling up the parameters of LLMs gives them amazing generalization capabilities compared to small pre-trained models, allowing them to achieve excellent performance in few-shot settings \cite{li2024rt, promptner}.
%
Paolini et al.~\cite{paolini2021structured} proposed the Translation between Augmented Natural Languages framework; Lu et al.~\cite{uie} proposed a text-to-structure generation framework; and Chen et al.~\cite{cp-ner} proposed collaborative domain-prefix tuning for NER. These methods have achieved state-of-the-art performance and demonstrated effectiveness in few-shot setting.
Despite the success of LLMs, they face challenges in training-free IE because of the difference between sequence labeling and text-generation models \cite{gutierrez2022thinking}. To overcome these limitations, GPT-NER \cite{gpt-ner} introduced a self-verification strategy, while GPT-RE \cite{gpt-re} enhanced task-aware representations and incorporates reasoning logic into enriched demonstrations. These methods effectively showcase how GPT can be leveraged for in-context learning. 
CODEIE \cite{codeie} and CodeKGC \cite{codekgc} showed that converting IE tasks into code generation tasks with code-style prompts and in-context examples leads to superior performance compared to NL-LLMs. This is because code-style prompts provide a more effective representation of structured output, enabling them to effectively handle the complex dependencies in natural language.

\subsection{Supervised Fine-tuning} \label{sec:full}
Using all training data to fine-tune LLMs is the most common and promising method~\cite{jiang2024toner, duan2024keyee, liu2024beyond, li2024recall, li2024empirical, cai2024improving, li2024relation, liu2024dr, guan2024langtopo,zha2024scaling,zhao2024comi,lin2024box}, which allows the model to capture the underlying structural patterns in the data, and generalize well to unseen samples. 
For example, DEEPSTRUCT~\cite{deepstruct} introduced structure pre-training on a collection of task-agnostic corpora to enhance the structural understanding of language models.
UniNER~\cite{universalner} explored targeted distillation and mission-focused instruction tuning to train student models for broad applications, such as NER.
GIELLM~\cite{giellm} fine-tuned LLMs using mixed datasets, which are collected to utilize the mutual reinforcement effect to enhance performance on multiple tasks.

\subsection{Summaries of Techniques}

Data augmentation~\cite{llmaaa,augure} is a widely explored direction due to its potential in enhancing model performance. LLMs possess extensive implicit knowledge and strong text generation capabilities, making them well-suited for data annotation tasks~\cite{chu2023survey}. However, while data augmentation can expand the training dataset and improve model generalization, they may also introduce noise. For example, knowledge retrieval methods can supply additional context about entities and relationships, enriching the extraction process. Nevertheless, the noise can detract from the overall quality of the extracted information~\cite{pgim,riveg}.  

On the other hand, designing effective prompts remains a significant challenge for leveraging LLMs like GPT-4~\cite{gpt4}. Although approaches such as QA dialogue and CoT~\cite{cot-er} strategies can enhance LLMs' IE capabilities, purely prompt-based methods still lag behind supervised fine-tuning with smaller models. Supervised fine-tuning~\cite{universalner,gner,instructuie}, including cross-domain and few-shot learning, generally yields better performance, which suggests that combining large-scale LLMs for data annotation with supervised fine-tuning using additional data can optimize performance and reduce manual annotation costs \cite{nuner,ye2024llm,evans2024astro}.

In summary, while various techniques for IE using LLMs offer distinct advantages, they also come with challenges. 
Thoughtfully combining these strategies can yield significant enhancements in IE tasks.

\begin{table*}[t]
\caption{The statistics of research in Specific Domain.}
\centering
\setlength\tabcolsep{5pt}  
\renewcommand{\arraystretch}{0.8} 
{ 
\small
\begin{tabular}{c|c|c|c|c}
\toprule
\textbf{Domain} &
  \textbf{Method} &
  \textbf{Task} &
  \textbf{Paradigm} &
  \textbf{Backbone}
  \\ \midrule
    \multirow{4}*{\textbf{Multimodal}}     
            & Cai et al.~\cite{cai2023context} & NER & ICL & GPT-3.5 \\
            & PGIM~\cite{pgim} & NER & DA & BLIP2, GPT-3.5 \\
            & RiVEG~\cite{riveg} & NER & DA & Vicuna, LLaMA2, GPT-3.5 \\
            & Chen et al.~\cite{chen2023chain} & NER, RE & DA & BLIP2, GPT-3.5, GPT-4 \\
    \midrule
    \multirow{3}*{\textbf{Multilingual}}
            & Meoni et al.~\cite{meoni2023large} & NER & DA & Text-davinci-003 \\
            & Naguib et al.~\cite{naguib2024few} & NER & ICL & - \\
            & Huang et al.~\cite{huang2022multilingual} & EE & CDL & mBART, mT5 \\

    \midrule
    \multirow{14}*{\textbf{Medical}}      
            & Bian et al.~\cite{bian2023inspire} & NER & DA & GPT-3.5 \\
            & Hu et al.~\cite{hu2023zero} & NER & ZS Pr & GPT-3.5, GPT-4 \\
            & Meoni et al.~\cite{meoni2023large} & NER & DA & Text-davinci-003 \\
            & Naguib et al.~\cite{naguib2024few} & NER & ICL & - \\
            & VANER~\cite{vaner} & NER & SFT & LLaMA2 \\
            & RT~\cite{li2024rt} & NER & ICL & GPT-4 \\
            & Munnangi et al.~\cite{munnangi2024fly} & NER & ZS Pr, ICL, FS FT & GPT-3.5, GPT-4, Claude-2, LLaMA2 \\
            & Monajatipoor et al.~\cite{monajatipoor2024llms} & NER & SFT, ICL & - \\
            & Hu et al.~\cite{hu2024improving} & NER & ZS Pr, ICL & GPT-3.5, GPT-4\\
            & Guti{\'e}rrez et al.~\cite{gutierrez2022thinking} & NER, RE & ICL & GPT-3 \\
            & GPT3+R~\cite{agrawal2022large} & NER, RE & - & Text-davinci-002 \\
            & Labrak et al.~\cite{labrak2023zero} & NER, RE & - & GPT-3.5, Flan-UL2, Tk-Insturct, Alpaca \\
            & Tang et al.~\cite{tang2023does} & NER, RE & DA & GPT-3.5 \\
            & DICE~\cite{dice} & EE & SFT & T5-Large \\
    \midrule
    \multirow{5}*{\textbf{Scientific}}
            & B{\"o}l{\"u}c{\"u} et al.~\cite{bolucu2023impact} & NER & ICL & GPT-3.5 \\
            & Dunn et al.~\cite{dunn2022structured} & NER, RE & SFT & GPT-3 \\
            & PolyIE~\cite{polyie} & NER, RE & ICL & GPT-3.5, GPT-4 \\
            & Foppiano et al.~\cite{foppiano2024mining} & NER, RE & ZS Pr, ICL, SFT & GPT-3.5, GPT-4 \\
            & Dagdelen et al.~\cite{dagdelen2024structured} & NER, RE & SFT & GPT-3, LLaMA2 \\
    \midrule
    \multirow{2}*{\textbf{Astronomical}}
            & Shao et al.~\cite{shao2024astronomical} & NER & ZS Pr & GPT-3.5, GPT-4, Claude-2, LLaMA2 \\ 
            & Evans et al.~\cite{evans2024astro} & NER & DA & GPT-3.5, GPT-4 \\

    \midrule
    \multirow{2}*{\textbf{Historical}}
            & Gonz{\'{a}}lez{-}Gallardo et al.~\cite{gonzalez2023yes} & NER  & ZS Pr & GPT-3.5 \\
            & CHisIEC~\cite{chisiec} & NER, RE & SFT, ICL & ChatGLM2, Alpaca2, GPT-3.5\\
    \midrule
    \multirow{3}*{\textbf{Legal}}
            & Nunes et al.~\cite{nunesout} & NER  & ICL & Sabia \\
            & Oliveira et al.~\cite{oliveira2024combining} & NER & DA & GPT-3 \\
            & Kwak et al.~\cite{KwakJFBMS23} & RE, EE & ICL & GPT-4 \\

\bottomrule
\end{tabular}
}

\label{tab:domain-table}

\end{table*}
\section{Applications on Specific Domains}\label{sec:domain}
It is non-ignorable that LLMs have tremendous potential for extracting information from some specific domains, such as mulitmodal~\cite{chen2023chain,pgim,cai2023context,riveg}, multilingual~\cite{huang2022multilingual,meoni2023large, naguib2024few}, medical \cite{dice,tang2023does,li2024rt,meoni2023large,hu2023zero,bian2023inspire,agrawal2022large,labrak2023zero,gutierrez2022thinking,vaner,munnangi2024fly,monajatipoor2024llms,hu2024improving,MOEMeetsLLMs,liu2024largedistilling}, scientific \cite{dunn2022structured,polyie,bolucu2023impact,foppiano2024mining,dagdelen2024structured}, astronomical~\cite{shao2024astronomical,evans2024astro}{, }historical~\cite{gonzalez2023yes,chisiec}, and legal~\cite{nunesout, oliveira2024combining, KwakJFBMS23}.
Additionally, we present statistical data in Table \ref{tab:domain-table}.

For instance, Chen et al.~\cite{chen2023chain} introduced a conditional prompt distillation method that enhances a model's reasoning ability by combining text-image pairs with chain-of-thought knowledge from LLMs, significantly improving performance in multimodal NER and multimodal RE. 
Tang et al.~\cite{tang2023does} explored the potential of LLMs in the field of clinical text mining and proposed a novel training approach, which leverages synthetic data to enhance performance and address privacy issues. 
Dunn et al.~\cite{dunn2022structured} presented a sequence-to-sequence approach by using GPT-3 for joint NER and RE from complex scientific text, demonstrating its effectiveness in extracting complex scientific knowledge in material chemistry.
Shao et al.~\cite{shao2024astronomical} explored the use of LLMs to extract astronomical knowledge entities from astrophysical journal articles. Conventional approaches encounter difficulties such as manual labor and limited generalizability. To address these issues, the authors proposed a prompting strategy that incorporates five prompt elements and eight combination prompt, aiming to specifically target celestial object identifiers and telescope names as the experimental objects of interest.
Gonz{\'{a}}lez et al.~\cite{gonzalez2023yes} examined the performance of ChatGPT in the NER task specifically on historical texts. The research not only compared ChatGPT with other state-of-the-art language model-based systems but also delved into the challenges encountered in this zero-shot setting. The findings shed light on the limitations of entity identification in historical texts, encompassing concerns related to annotation guidelines, entity complexity, code-switching, and the specificity of prompts.

\section{Evaluation \& Analysis}\label{sec:eval}
Despite the great success of LLMs in various natural language processing tasks \cite{wang2024llm4msr,10506571}, their performance in the field of information extraction still have room for improvement \cite{han2023information}. 
To alleviate this problem, recent research has explored the capabilities of LLMs with respect to the major subtasks of IE, i.e., NER \cite{xie2023empirical, naguib2024few}, RE \cite{wadhwa2023revisiting, yuan2023zero}, and EE \cite{gao2023exploring}.
Considering the superior reasoning capabilities of LLMs, Xie et al.~\cite{xie2023empirical} proposed four reasoning strategies for NER, which are designed to simulate ChatGPT's potential on zero-shot NER.
Wadhwa et al.~\cite{wadhwa2023revisiting} explored the use of LLMs for RE and found that few-shot prompting with GPT-3 achieves near SOTA performance, while Flan-T5 can be improved with chain-of-thought style explanations generated via GPT-3.
For EE task, Gao et al.~\cite{gao2023exploring} showed that ChatGPT still struggles with it due to the need for complex instructions and a lack of robustness.

Along this line, some researchers performed a more comprehensive analysis of LLMs by evaluating multiple IE subtasks simultaneously.
Li et al.~\cite{li2023evaluating} evaluated ChatGPT's overall ability on IE, including performance, explainability, calibration, and faithfulness. They found that ChatGPT mostly performs worse than BERT-based models in the standard IE setting, but excellently in the OpenIE setting.
Furthermore, Han et al.~\cite{han2023information} introduced a soft-matching strategy for a more precise evaluation and identified ``unannotated spans'' as the predominant error type, highlighting potential issues with data annotation quality.
\section{Future Directions}\label{sec:future}

The development of LLMs for generative IE is still in its early stages, and there are numerous opportunities for improvement.

\noindent
\textbf{Universal IE.} Previous generative IE methods and benchmarks are often tailored for specific domains or tasks, limiting their generalizability \cite{get}. Although some unified methods \cite{uie} using LLMs have been proposed recently, they still suffer from certain limitations (e.g., long context input, and misalignment of structured output). Therefore, further development of universal IE frameworks that can adapt flexibly to different domains and tasks is a promising research direction (such as integrating the insights of task-specific models to assist in constructing universal models).

\noindent
\textbf{Low-Resource IE.} The generative IE system with LLMs still encounters challenges in resource-limited scenarios \cite{li2023evaluating}.
There is a need for further exploration of in-context learning of LLMs, particularly in terms of improving the selection of examples. Future research should prioritize the development of robust cross-domain learning techniques \cite{instructuie}, such as domain adaptation or multi-task learning, to leverage knowledge from resource-rich domains. Additionally, efficient data annotation strategies with LLMs should also be explored.

\noindent
\textbf{Prompt Design for IE.} 
Designing effective instructions is considered to have a significant impact on the performance of LLMs \cite{qiao2022reasoning,yin2023survey}.
One aspect of prompt design is to build input and output pairs that can better align with pre-training stage of LLMs (e.g., code generation) \cite{code4uie}. Another aspect is optimizing the prompt for better model understanding and reasoning (e.g., Chain-of-Thought) \cite{li2023revisiting}, by encouraging LLMs to make logical inferences or explainable generation. 
Additionally, researchers can explore interactive prompt design (such as multi-turn QA) \cite{qa4re}, where LLMs can iteratively refine or provide feedback on the generated extractions automatically.

\noindent
\textbf{Open IE.} 
Open IE settings present greater challenges for IE models, as they do not provide a candidate label set and rely solely on the models' ability to comprehend the task. LLMs, with their knowledge and understanding abilities, have significant advantages in some Open IE tasks \cite{universalner}. However, there are still instances of poor performance in more challenging tasks \cite{qi2023preserving}, which require further exploration by researchers.

\section{Benchmarks \& Backbones}\label{sec:Benchmarks}
\subsection{Representative Datasets.} 
\begin{table}[h]
\caption{A summary of some representative IE datasets.}
\centering
\setlength\tabcolsep{3pt}  
\resizebox{\linewidth}{!}{
\begin{tabular}{c|c}
\toprule
  \textbf{Dataset} &
  \textbf{Summary}  \\ \midrule
CoNLL03 \cite{conll03} & \makecell[c]{\textbf{Dataset scope: NER}; \\1,393 English news articles from Reuters; \\ 909 German news articles; \\4 annotated entity types.} \\ \midrule
CoNLL04 \cite{conll04} & \makecell[c]{\textbf{Dataset scope: RE}; \\ entity-relation triples from news sentences; \\4 entity types; \\5 relation types.} \\ \midrule
ACE05 \cite{ace05} & \makecell[c]{\textbf{Dataset scope: NER, RE and EE}; \\various text types and genres; \\7 entity types; 7 relation types; \\33 event types and 22 argument roles.} \\
\bottomrule
\end{tabular}
}
\label{tab:summary-table}
\end{table}
In this section, we introduce representative datasets of NER, RE and EE respectively, and show brief summary of each dataset in the Table \ref{tab:summary-table} to help readers better understand these tasks.

\noindent
\textbf{CoNLL03.} The CoNLL03 \cite{conll03} is a representative dataset for NER, including 1,393 news articles in English and 909 news articles in German. The English portion of the corpus was sourced from the Shared Task dataset curated by Reuters. This dataset encompasses annotations for four distinct entity types: PER (person), LOC (location), ORG (organization), and MISC (including all other types of entities).

\noindent
\textbf{CoNLL04.} The CoNLL04 \cite{conll04} is a well-known benchmark dataset for RE tasks, comprising sentences extracted from news articles that each contain at least one entity-relation triple. It has four kinds of entities (PER, ORG, LOC, OTH) and five kinds of relations (Kill, Work For, Live In, OrgBased In, Located In).

\noindent
\textbf{ACE05.} Automatic Content Extraction 05 \cite{ace05} is widely recognized and utilized for IE tasks.
It serves as a valuable resource to assess the efficacy of automated systems in extracting structured information from diverse textual sources, encompassing news articles, interviews, reports, etc. Moreover, this dataset covers a broad range of genres including politics, economics, sports, among others.
Specifically for the EE task within ACE05, it comprises 599 news documents that encapsulate 33 distinct event types and 22 argument roles.
\begin{table*}[htbp]
\caption{Statistics of common datasets for information extraction. $*$ denotes the dataset is multimodal. \# refers to the number of categories or sentences. The data in the table is partially referenced from InstructUIE \cite{instructie}.}
\centering
\setlength\tabcolsep{11pt}  
\renewcommand{\arraystretch}{0.8} 
{ 
\small
\begin{tabular}{c|c|c|c|c|c|c}
\toprule
\textbf{Task} &
  \textbf{Dataset} &
  \textbf{Domain} &
  \textbf{\#Class} &
  \textbf{\#Train} &
  \textbf{\#Val} &
  \textbf{\#Test} 
  \\ \midrule
    \multirow{32}*{\textbf{NER}}     
            &   ACE04~\cite{ace04}   &   News     &   7   &   6,202    &   745     &   812     \\
            &   ACE05~\cite{ace05}   &   News     &   7   &   7,299    &   971     &   1,060    \\
            &   BC5CDR~\cite{bc5cdr} &  Biomedical  &   2   &   4,560    &   4,581    &   4,797    \\
            &   Broad Twitter Corpus~\cite{btc} &   Social Media  &  3  &  6,338  &  1,001  &  2,000   \\         
            &   CADEC~\cite{cadec}   &   Biomedical   &   1   &   5,340   &   1,097   &  1,160    \\
            &   CoNLL03~\cite{conll03} &   News     &   4   &   14,041   &   3,250    &   3,453    \\
            &   CoNLLpp~\cite{conllpp} &   News     &   4   &   14,041   &   3,250    &   3,453    \\
            &   CrossNER-AI~\cite{crossner} &  Artificial Intelligence    &   14  &   100 &   350 &   431 \\
            &   CrossNER-Literature~\cite{crossner} &  Literary    &   12  &   100 &   400 &   416 \\
            &   CrossNER-Music~\cite{crossner} &  Musical    &   13  &   100 &   380 &   465 \\
            &   CrossNER-Politics~\cite{crossner} &  Political    &   9  &   199 &   540 &   650 \\
            &   CrossNER-Science~\cite{crossner} &  Scientific    &   17  &   200 &   450 &   543 \\
            &   FabNER~\cite{fabner}    &   Scientific  &   12  &   9,435    &   2,182    &   2,064    \\
            &   Few-NERD~\cite{fewnerd} &   General &   66  &   131,767  &   18,824   &   37,468   \\
            &   FindVehicle~\cite{findvehicle} &   Traffic   &   21   &   21,565   &   20,777   &   20,777   \\
            &   GENIA~\cite{genia}   &   Biomedical &   5   &   15,023   &   1,669    &   1,854    \\
            &   HarveyNER~\cite{harveyner} & Social Media &     4   &   3,967    &   1,301    &   1,303    \\
            &   MIT-Movie~\cite{mit-dataset} &    Social Media    &   12  &   9,774    &   2,442    &   2,442    \\
            &   MIT-Restaurant~\cite{mit-dataset} &   Social Media   &   8  &   7,659    &   1,520   &   1,520    \\
            &   MultiNERD~\cite{multinerd}  &   Wikipedia   &   16   &   134,144   &   10,000   &  10,000     \\
            &   NCBI~\cite{ncbi} &  Biomedical  &   4   &   5,432    &   923    &   940    \\
            &   OntoNotes 5.0~\cite{ontonotes}   &   General   &   18  &   59,924   &   8,528    &   8,262    \\
            &   ShARe13~\cite{share13} &   Biomedical   &   1   &   8,508   &   12,050   &  9,009    \\
            &   ShARe14~\cite{share14} &   Biomedical   &   1   &   17,404   &   1,360   &  15,850    \\
            &   SNAP$^*$~\cite{twitter2017_snap} &  Social Media  &   4   &   4,290    &   1,432    &   1,459    \\
            &   TTC~\cite{ttc} & Social Meida &   3   &   10,000   &   500   &   1,500   \\   
            &   Tweebank-NER~\cite{tweebank-ner} &  Social Media  &   4   &   1,639    &   710     &   1,201   \\
            &   Twitter2015$^*$~\cite{twitter2015} &  Social Media  &   4   &   4,000   &  1,000   &   3,357    \\
            &   Twitter2017$^*$~\cite{twitter2017_snap} &  Social Media  &   4   &  3,373  &  723  &   723    \\
            &   TwitterNER7~\cite{twitterner7} &  Social Media  &   7   &  7,111  &  886  &   576    \\
            &   WikiDiverse$^*$~\cite{wikidiverse} &  News    &   13   &     6,312    &   755    &   757    \\
            &   WNUT2017~\cite{wnut2017}    &   Social Media &    6   &   3,394   &   1,009   &  1,287    \\
    \midrule
    \multirow{11}*{\textbf{RE}}      
            &   ACE05~\cite{ace05}   &   News     &   7   &   10,051   &   2,420    &   2,050    \\
            &   ADE~\cite{ade} &   Biomedical       &   1   &   3,417    &   427     &   428    \\
            &   CoNLL04~\cite{conll04} &   News    &   5   &   922     &   231     &   288    \\
            &   DocRED~\cite{docred} &   Wikipedia  &   96   &   3,008     &   300     &   700    \\
            &   MNRE$^*$~\cite{mnre} &  Social Media    &   23   &   12,247    &   1,624    &   1,614    \\
            &   NYT~\cite{nyt} &   News     &   24   &   56,196     &   5,000     &   5,000    \\
            &   Re-TACRED~\cite{re-tacred}  &   News     &   40   &   58,465     &   19,584    &   13,418    \\
            &   SciERC~\cite{scierc}  &   Scientific   &   7   &   1,366     &   187    &   397    \\
            &   SemEval2010~\cite{semeval}  &   General   &   19   &   6,507     &   1,493    &   2,717    \\ 
            &   TACRED~\cite{tacred}  &   News     &   42   &   68,124     &   22,631    &   15,509    \\
            &   TACREV~\cite{tacrev}  &   News     &   42   &   68,124     &   22,631    &   15,509    \\
    \midrule
    \multirow{7}*{\textbf{EE}}      
            &   ACE05~\cite{ace05}   &   News     &   33/22   &   17,172   &   923    &   832    \\
            &   CASIE~\cite{casie}   &   Cybersecurity     &   5/26   &   11,189   &   1,778    &   3,208    \\
            &   GENIA11~\cite{genia11}   &   Biomedical  &   9/11   &   8,730   &   1,091   &    1,092   \\
            &   GENIA13~\cite{genia13}   &   Biomedical  &   13/7   &   4,000   &   500   &    500   \\
            &   PHEE~\cite{phee}   &   Biomedical  &   2/16   &   2,898   &   961   &    968   \\
            &   RAMS~\cite{rams}   &   News     &   139/65   &   7,329   &   924    &   871    \\
            &   WikiEvents~\cite{wikievents}   &   Wikipedia &   50/59   &   5,262   &   378    &   492    \\
            
\bottomrule
\end{tabular}
}
\label{tab:datasets-table}

\end{table*}

\begin{table*}[t]
\caption{The common backbones for generative information extraction. We mark the commonly used base and large versions for better reference.}
\centering
\setlength\tabcolsep{7pt}  
\renewcommand{\arraystretch}{0.6} 
{ 
\begin{tabular}{@{}c|cccccc@{}}
\toprule
  \textbf{Series}  &
\textbf{Model} &
  \textbf{Size} &
  \textbf{\makecell[c]{Base\\Model}} &
  \textbf{\makecell[c]{Open\\Source}} &
  \textbf{\makecell[c]{Instruction\\Tuning}} &  
  \textbf{RLHF}
  \\ 
\midrule
BART  &   BART~\cite{bart}    &  \makecell[c]{140M (base),\\ 400M (large)}  &   -    &   $\surd$ &   -   &   -       \\
\midrule
\multirow{3}*{T5} 
    &   T5~\cite{t5}     &   \makecell[c]{60M, 220M (base), \\770M (large), 3B, 11B}    &   -      &   $\surd$ &   -       &   -     \\
    &   mT5~\cite{mt5}    &   \makecell[c]{300M, 580M (base), \\1.2B (large), 3.7B, 13B}    &   -   &   $\surd$ &   -       &   -     \\
    &   Flan-T5~\cite{flan-t5}    &   \makecell[c]{80M, 250M (base), \\780M (large), 3B, 11B}  &   T5   &   $\surd$ &   $\surd$ &   -     \\ 
\midrule
\multirow{2}*{GLM}
    &   GLM~\cite{glm}             &   \makecell[c]{110M (base), \\335M (large), \\410M, 515M, 2B, 10B}  &  -     &   $\surd$ &   -       &   -       \\
    &   ChatGLM series         &   6B    &  GLM   &   $\surd$ &   $\surd$   &   $\surd$       \\
\midrule
\multirow{7}*{LLaMA}
    &   LLaMA~\cite{llama}           &   7B, 13B, 33B, 65B &  -     &   $\surd$ &   -       &   -       \\
    &   Alpaca~\cite{alpaca}          &   7B, 13B           &  LLaMA &   $\surd$ &   $\surd$ &   -       \\
    &   Vicuna~\cite{vicuna}          &   7B, 13B           &  LLaMA &   $\surd$ &   $\surd$ &   -       \\
    &   LLaMA2~\cite{llama2}          &   7B, 13B, 70B      &  -     &   $\surd$ &   -       &   -       \\
    &   LLaMA2-chat~\cite{llama2}     &   7B, 13B, 70B      &  LLaMA2     &   $\surd$ &  $\surd$  &   $\surd$    \\   
    &   Code-LLaMA~\cite{code-llama}      &   7B, 13B, 34B      &  LLaMA2     &   $\surd$ &   -       &   -       \\
    &   LLaMA3 series       &   8B, 70B, 405B  &  -  &  $\surd$ &  $\surd$  &   $\surd$    \\   
\midrule
\multirow{8}*{GPT}
    &   GPT-2~\cite{gpt2}           &   \makecell[c]{117M, 345M, 762M, \\1.5B}       &   -           &   $\surd$  &   -       &   -       \\
    &   GPT-3~\cite{gpt3}           &   175B    &   -           &   -       &   -       &   -       \\
    &   GPT-J~\cite{gpt-j}           &   6B      &   GPT-3       &   $\surd$ &   -       &   -       \\
    &   Code-davinci-002~\cite{instructgpt} &   -       &   GPT-3       &   -       &   $\surd$ &   -       \\
    &   Text-davinci-002~\cite{instructgpt} &   -       &   GPT-3       &   -       &   $\surd$ &   -       \\
    &   Text-davinci-003~\cite{instructgpt} &   -       &   GPT-3       &   -       &   $\surd$ &   $\surd$ \\
    &   GPT-3.5-turbo series~\cite{chatgpt}   &   -       &   -           &   -       &   $\surd$ &   $\surd$ \\
    &   GPT-4 series~\cite{gpt4}           &   -       &   -           &   -       &   $\surd$ &   $\surd$ \\

\bottomrule
\end{tabular}

}
\label{tab:backbones-table}

\end{table*}

\subsection{Benchmarks.} 
As shown in Table \ref{tab:datasets-table}, we compiled a comprehensive collection of benchmarks covering various domains and tasks, to provide researchers with a valuable resource that they can query and reference as needed. Moreover, we also summarized the download links for each dataset in our open source repository (\href{https://github.com/quqxui/Awesome-LLM4IE-Papers}{LLM4IE repository}).

\subsection{Backbones.} \label{sec:Backbones}
We briefly describe some backbones that are commonly used in the field of generative information extraction, which is shown in Table \ref{tab:backbones-table}.

\section{Conclusion}
In this survey, We first introduced the subtasks of IE and discussed some universal frameworks aiming to unify all IE tasks with LLMs. Additional theoretical and experimental analysis provided insightful exploration for these methods.
Then we delved into different IE techniques that apply LLMs for IE and demonstrate their potential for extracting information in specific domains.
Finally, we analyzed the current challenges and presented potential future directions. We hope this survey can provide a valuable resource for researchers to explore more efficient utilization of LLMs for IE.

\begin{acknowledgement}
This work was supported in part by the grants from National Natural Science Foundation of China (No.62222213, 62072423). Additionally, this research was partially supported by Research Impact Fund (No.R1015-23), APRC - CityU New Research Initiatives (No.9610565, Start-up Grant for New Faculty of CityU), CityU - HKIDS Early Career Research Grant (No.9360163), Hong Kong ITC Innovation and Technology Fund Midstream Research Programme for Universities Project (No.ITS/034/22MS), Hong Kong Environmental and Conservation Fund (No. 88/2022), and SIRG - CityU Strategic Interdisciplinary Research Grant (No.7020046), Huawei (Huawei Innovation Research Program), Tencent (CCF-Tencent Open Fund, Tencent Rhino-Bird Focused Research Program), Ant Group (CCF-Ant Research Fund, Ant Group Research Fund), Alibaba (CCF-Alimama Tech Kangaroo Fund (No. 2024002)), CCF-BaiChuan-Ebtech Foundation Model Fund, and Kuaishou.
\end{acknowledgement}

\bibliographystyle{fcs}
\bibliography{template}

\begin{biography}{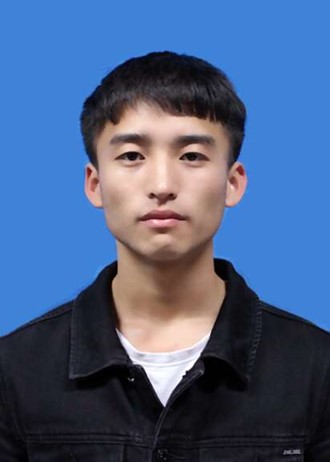}
Derong Xu is currently a joint PhD student at University of Science and Technology of China and City University of Hong Kong. His research interests focus on Multimodal Knowledge graph and large language models.
\end{biography}


\begin{biography}{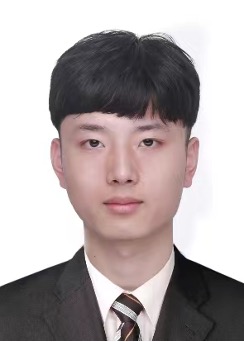}
Wei Chen is now a doctoral student at the University of Science and Technology of China. His research interests include data mining, information extraction, and large language models.
\end{biography}
\begin{biography}{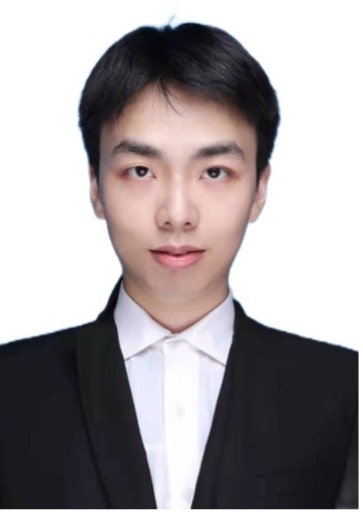}
Wenjun Peng received his Master's degree in the School of Computer Science and Technology at University of Science and Technology of China (USTC). He obtained his B.E. degree from Sichuan University in 2021. His main research interests include data mining, multimodal learning and person re-ID.
\end{biography}
\begin{biography}{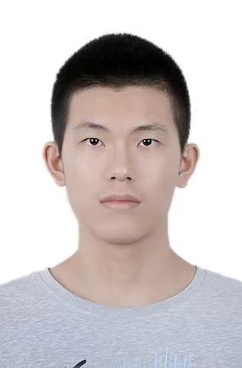}
Chao Zhang received the B.E. degree in software engineering from Shandong University in 2022. He is currently pursuing a joint Ph.D. degree at the University of Science and Technology of China and City University of Hong Kong. His research interests include data mining, multimodal learning, and large language models.
\end{biography}
\begin{biography}{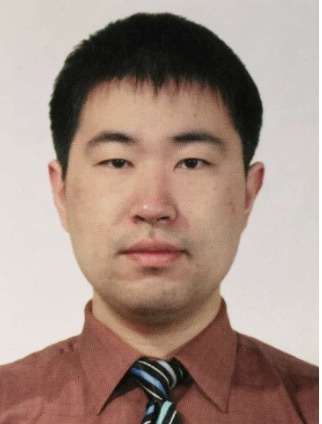}
Tong Xu is currently working as a Professor at University of Science and Technology of China (USTC), Hefei, China. He has authored more than 100 top-tier journal and conference papers in related fields, including TKDE, TMC, TMM, TOMM, KDD, SIGIR, WWW, ACM MM, etc. He was the recipient of the Best Paper Award of KSEM 2020, and the Area Chair Award for NLP Application Track of ACL 2023.
\end{biography}
\begin{biography}{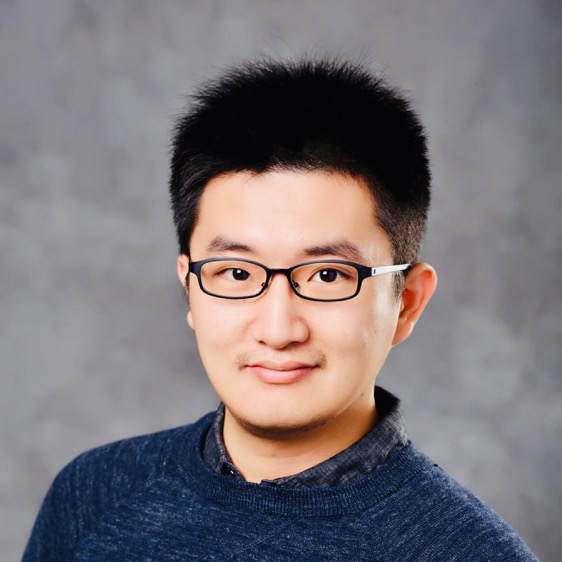}
Xiangyu Zhao is an assistant professor of the school of data science at City University of Hong Kong (CityU). His current research interests include data mining and machine learning. He has published more than 100 papers in top conferences and journals. His research has been awarded ICDM’22 and ICDM’21 Best-ranked Papers, Global Top 100 Chinese New Stars in AI, Huawei Innovation Research Program, CCF-Tencent Open Fund (twice), CCF-Ant Research Fund, Ant Group Research Fund, Tencent Focused Research Fund, and nomination for Joint AAAI/ACM SIGAI Doctoral Dissertation Award. He serves as top data science conference (senior) program committee members and session chairs, and journal guest editors and reviewers.
\end{biography}
\begin{biography}{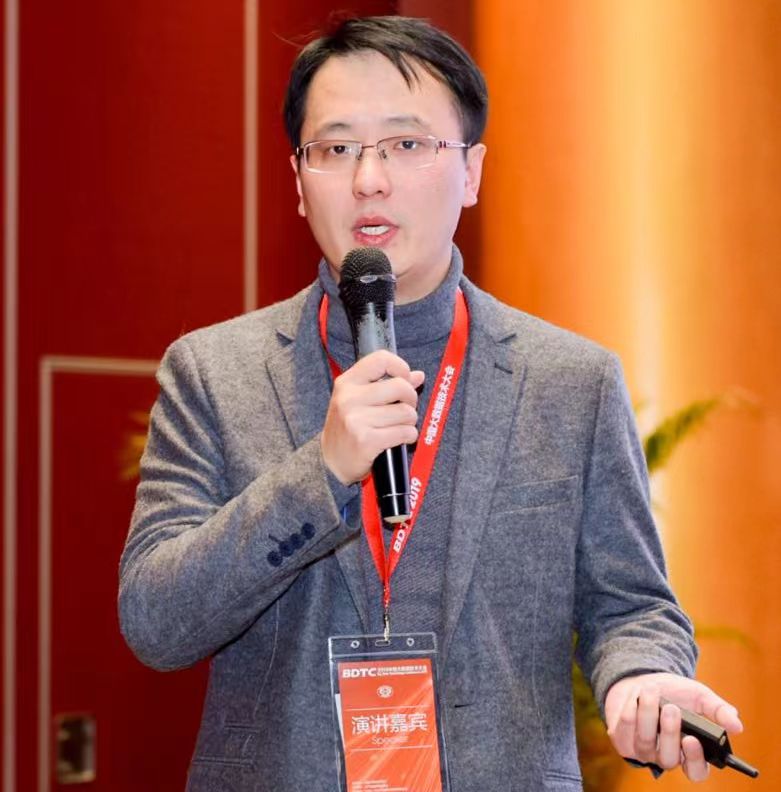}
Xian Wu is now a Principal Researcher in Tencent. Before joining Tencent, he worked as a Senior Scientist Manager and a Staff Researcher in Microsoft and IBM Research. Xian Wu received his PhD degree from Shanghai Jiao Tong University. His research interests includes Medical AI, Natural Language Processing and Multi-Modal modeling. Xian Wu has published papers in Nature Computational Science, NPJ digital medicine, T-PAMI, CVPR, NeurIPS, ACL, WWW, KDD, AAAI, IJCAI etc. He also served as PC member of BMJ, T-PAMI, TKDE, TKDD, TOIS, TIST, CVPR, ICCV, AAAI etc.
\end{biography}
\begin{biography}{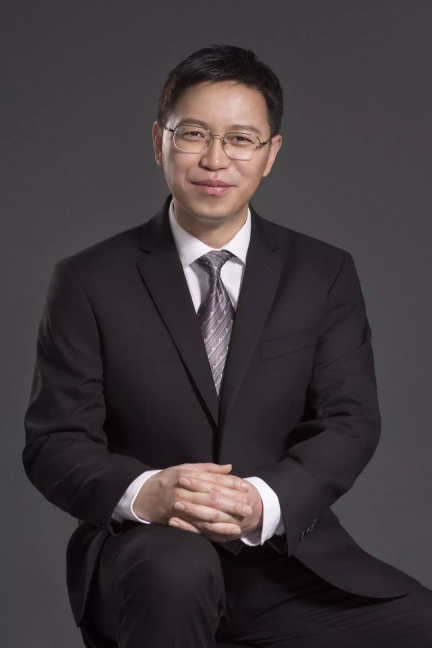}
Yefeng Zheng received B.E. and M.E. degrees from Tsinghua University, Beijing, China in 1998 and 2001, respectively, and a Ph.D. degree from University of Maryland, College Park, USA in 2005. After graduation, he worked at Siemens Corporate Research in Princeton, New Jersey, USA on medical image analysis before joining Tencent in Shenzhen, China in 2018. He is now Distinguished Scientist and Director of Tencent Jarvis Research Center, leading the company's initiative on medical artificial intelligence. He has published 300+ papers and invented 80+ US patents. His work has been cited more than 22,000 times with h-index of 74. He is a fellow of IEEE, a fellow of AIMBE, and an Associate Editor of IEEE Transactions on Medical Imaging.
\end{biography}
\begin{biography}{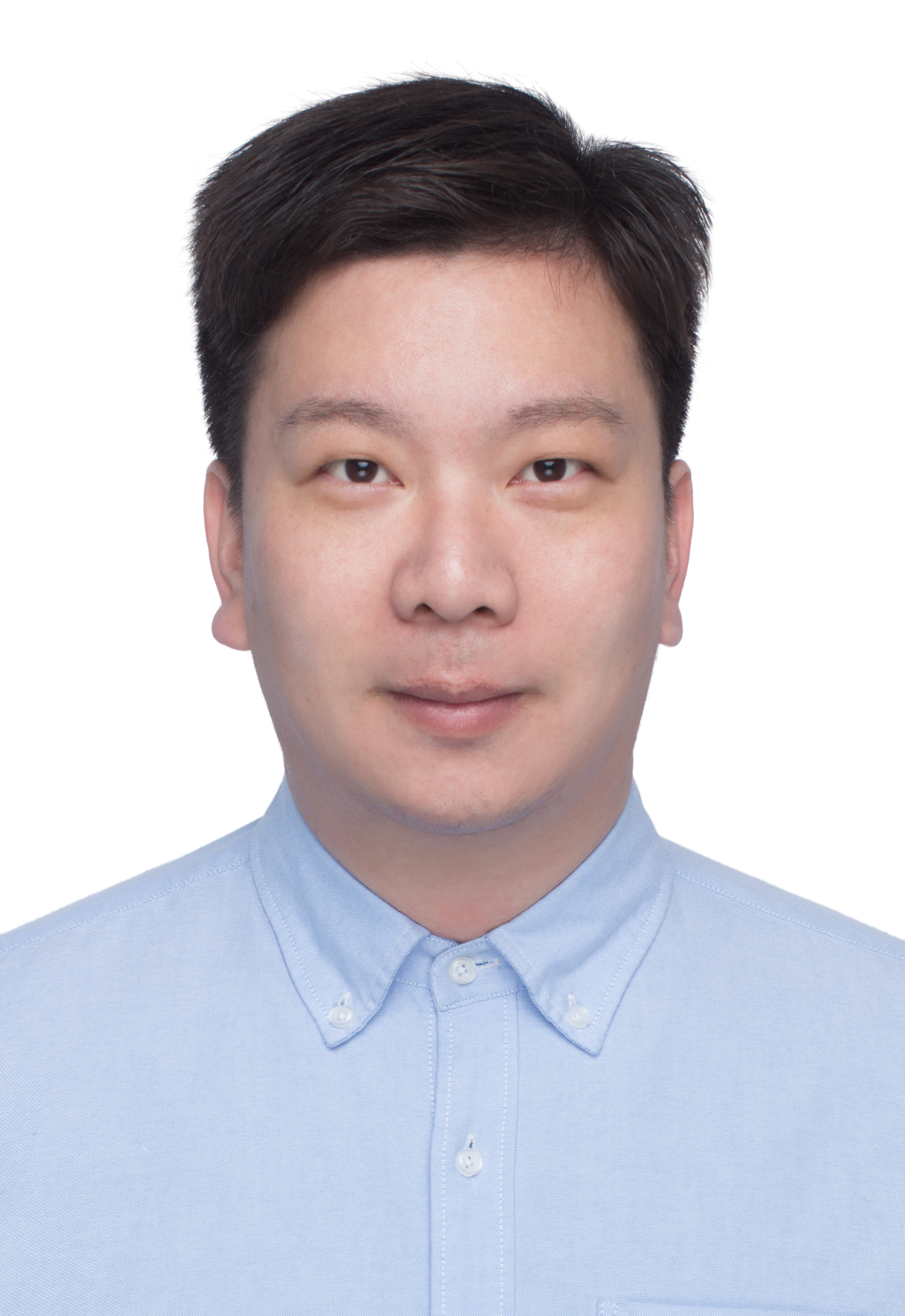}
Yang Wang is currently working as an Engineer at Anhui Conch Information Technology Engineering Co., Ltd., Wuhu, China. He has more than 10 years of IT project implementation experience in building materials industry, applied for 11 invention patents, published 2 technological papers, and participated in 3 large-scale national science and technology projects.
\end{biography}
\begin{biography}{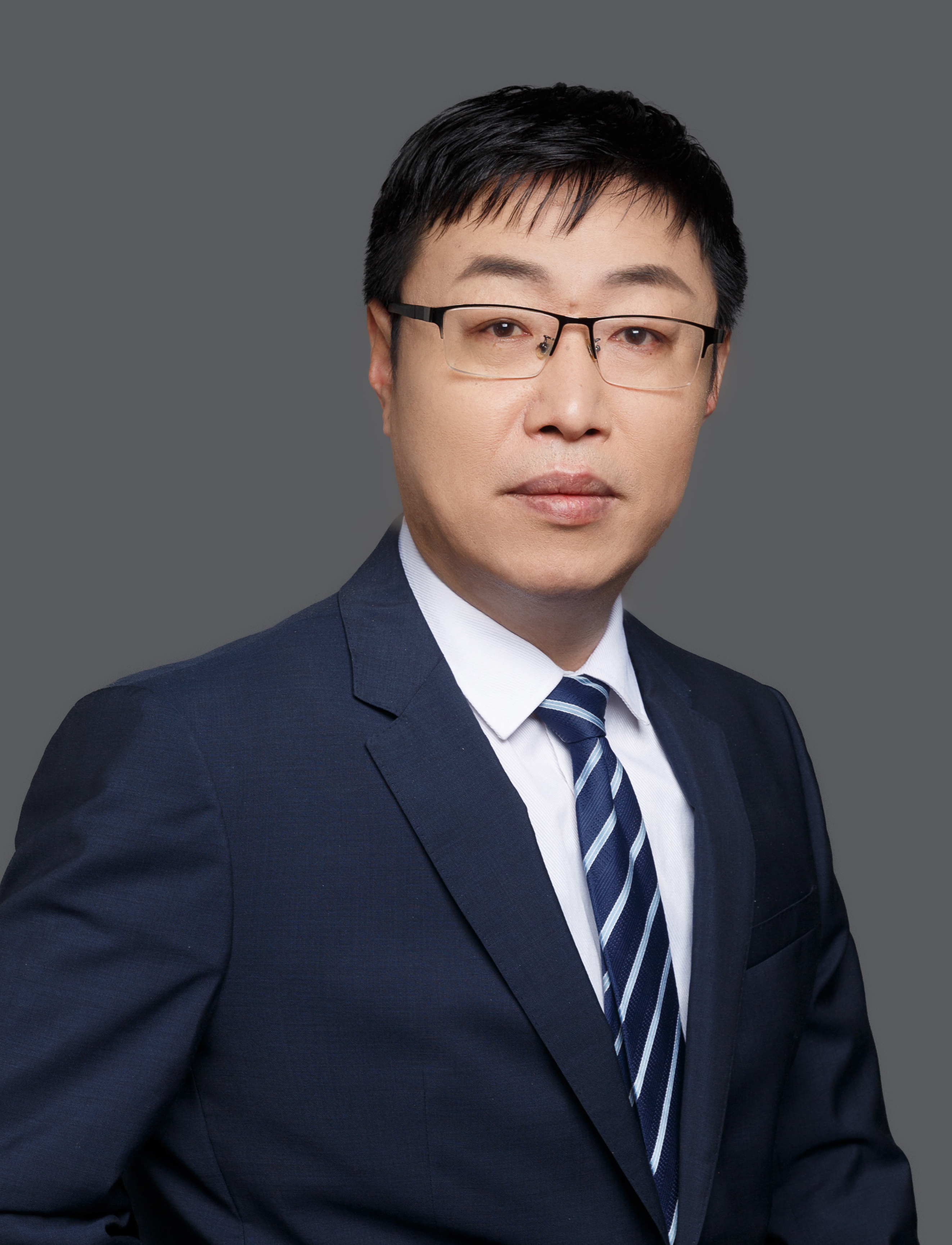}
Enhong Chen (CCF Fellow, IEEE Fellow) is a professor of University of Science and Technology of China (USTC). His general area of research includes data mining and machine learning, social network analysis, and recommender systems. He has published more than 300 papers in refereed conferences and journals, including Nature Communications, IEEE/ACM Transactions, KDD, NIPS, IJCAI and AAAI, etc. He was on program committees of numerous conferences including KDD, ICDM, and SDM. He received the Best Application Paper Award on KDD-2008, the Best Research Paper Award on ICDM-2011, and the Best of SDM-2015. His research is supported by the National Science Foundation for Distinguished Young Scholars of China.
\end{biography}


\end{document}